\documentclass[conference,compsoc]{IEEEtran}
\IEEEoverridecommandlockouts
\usepackage{tikz}
\usepackage{amsmath}
\usepackage{booktabs}
\usepackage{multirow}
\usepackage{algorithm}
\usepackage{algpseudocode}
\usepackage{xcolor}
\usepackage{hyperref}
\usepackage{amsmath}
\usepackage{amssymb}
\usepackage{caption}
\usepackage{fontawesome5}  
\usepackage{url} 

\usepackage{filecontents}
\newcommand{\halfblackwhitecircle}{
    \begin{tikzpicture}
        \draw[fill=black] (0,0) circle (0.8mm);
        \fill[white] (0,0) -- (0:0.8mm) arc[start angle=0, end angle=180, radius=0.8mm] -- cycle;
    \end{tikzpicture}
}
\newcommand{\whitecircle}{
    \begin{tikzpicture}
        \draw[fill=white] (0,0) circle (0.8mm);
    \end{tikzpicture}
}

\newcommand{\blackcircle}{
    \begin{tikzpicture}
        \draw[fill=black] (0,0) circle (0.8mm);
    \end{tikzpicture}
}

\ifCLASSOPTIONcompsoc
  \usepackage[nocompress]{cite}
\else
  \usepackage{cite}
\fi
\ifCLASSINFOpdf
\else
\fi
\begin{document}
%
\title{Not All Edges are Equally Robust: Evaluating the Robustness of Ranking-Based Federated Learning}

\author{
    \IEEEauthorblockN{
        Zirui Gong\textsuperscript{1}, 
        Yanjun Zhang\textsuperscript{2}, 
        Leo Yu Zhang\textsuperscript{\faEnvelope 1}, 
        Zhaoxi Zhang\textsuperscript{2,1}, 
        Yong Xiang\textsuperscript{3}, 
        Shirui Pan\textsuperscript{1}
    }
    \IEEEauthorblockA{\textsuperscript{1} Griffith University} 
    \IEEEauthorblockA{\textsuperscript{2} University of Technology Sydney}
    \IEEEauthorblockA{\textsuperscript{3} Deakin University}
    \thanks{\textsuperscript{\faEnvelope} Correspondence to Leo Yu Zhang \href{mailto:leo.zhang@griffith.edu.au}{(leo.zhang@griffith.edu.au).}}

}


%


\maketitle
\begin{abstract}
Federated Ranking Learning (FRL) is a state-of-the-art FL framework that stands out for its communication efficiency and resilience to poisoning attacks. It diverges from the traditional FL framework in two ways: 1) it leverages discrete rankings instead of gradient updates, significantly reducing communication costs and limiting the potential space for malicious updates, and 2) it uses majority voting on the server side to establish the global ranking, ensuring that individual updates have minimal influence since each client contributes only a single vote. These features enhance the system's scalability and position FRL as a promising paradigm for FL training.

However, our analysis reveals that FRL is not inherently robust, as certain edges are particularly vulnerable to poisoning attacks. Through a theoretical investigation, we prove the existence of these vulnerable edges and establish a lower bound and an upper bound for identifying them in each layer.
Based on this finding, we introduce a novel local model poisoning attack against FRL, namely \underline{V}ulnerable \underline{E}dge \underline{M}anipulation (VEM) attack. The VEM attack focuses on identifying and perturbing the most vulnerable edges in each layer and leveraging an optimization-based approach to maximize the attack's impact.
Through extensive experiments on benchmark datasets, we demonstrate that our attack achieves an overall 53.23\% attack impact and is 3.7$\times$ more impactful than existing methods. Our findings highlight significant vulnerabilities in ranking-based FL systems and underline the urgency for the development of new robust FL frameworks.
\end{abstract}


%
\IEEEpeerreviewmaketitle

\section{Introduction}
Federated Learning (FL) \cite{mcmahan2017communication} is a decentralized approach to machine learning where multiple clients collaborate to train a shared global model without sharing their private data. 
In FL, each client retains a private dataset and trains a local model using this data. After local training, clients send only the model updates to a central server for aggregation. After the aggregation, the server broadcasts the updated global model back to the clients for the next training cycle. FL has gained popularity and has been widely adopted by the industry as it allows for the training of machine learning models with vast amounts of data without exposing private raw data to the server~\cite{yu2016iprivacy,aouedi2022handling}.


\begin{figure}
    \centering
    \includegraphics[width=0.95\columnwidth]{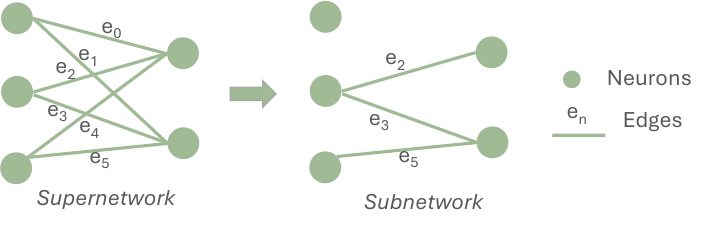}
    \caption{Example of Lottery Tickets Hypothesis (LTH).}
    \label{fig:lth}
\end{figure}
However, there are two major factors that limit the practicality of FL in real-world applications. The primary issue is communication bottleneck \cite{bernstein2018signsgd,alistarh2018convergence}. Traditional FL frameworks require frequent uploads and downloads of 32-bit weight parameters, which significantly constrains scalability. This issue is further amplified with the development of larger and deeper neural networks. This high communication overhead can also delay training and make the deployment of large-scale FL systems inefficient.
Another significant issue is the security and robustness concerns \cite{biggio2012poisoning,kurakin2016adversarial,munoz2017towards,jagielski2018manipulating,fang2020local,shejwalkar2021manipulating,bagdasaryan2020backdoor,wang2020attack,li20233dfed}. The decentralized nature of FL makes it vulnerable to client-side poisoning attacks. Malicious clients can manipulate their local updates to degrade the global model's performance. To address these two challenges, a recent work Federated Ranking Learning (FRL) was proposed as a promising solution \cite{mozaffari2023every}.

FRL is founded on the Lottery Tickets Hypothesis (LTH) \cite{frankle2018lottery}, which posits that within a randomly initialized, dense neural network (the \textit{supernetwork}), there exists a sparse \textit{subnetwork} that can achieve the same test accuracy as the fully trained \textit{supernetwork} (as illustrated in Fig. \ref{fig:lth}). Building on this concept, FRL facilitates collaboration among clients to identify a \textit{subnetwork} that performs well across all clients. 
In this process, clients evaluate the importance of each edge $e$ (the connections between neurons) based on their local datasets, assign a ranking to these edges, and send the ranking to the server. The server then aggregates these rankings using a majority voting mechanism to produce a global ranking. Finally, the top k\% edges with the highest ranking are selected to form the final \textit{subnetwork}.
Unlike the traditional FL, where the server aggregates 32-bit `model updates' from trusted clients and adjusts the global model's weights.
In FRL, the clients send their updates in the form of edge `rankings', which are permutations of integers within the range $[1, n]$, and $n$ is the number of edges in each layer.
This rule significantly reduces the upload and download communication cost and enhances its scalability to large networks. As for the security issue, the discrete update narrows the potential space for malicious updates from an infinite range to $n!$, effectively bounding the adversary's damage within a defined budget and thereby substantially mitigating the impact of potential attacks. On the server side, FRL utilizes majority voting to determine the global ranking. This approach prevents malicious clients from making significant adversarial modifications to the global model, as each client only has a single vote. 
These two designs restrict the adversary's capacity to scale or optimize the malicious updates in the continuous space, making existing model poisoning attacks (MPA) \cite{biggio2012poisoning,kurakin2016adversarial,munoz2017towards,jagielski2018manipulating,fang2020local,shejwalkar2021manipulating,bagdasaryan2020backdoor,wang2020attack,li20233dfed,zhang2023denial} ineffective against the FRL system.
Given the above advantages, it is valuable to further evaluate this ranking-based FL framework and determine whether this promising system can be reliably and safely deployed in real-world applications. 

\textbf{Our work:} 
In this work, we conduct the first thorough analysis of the robustness of this promising FL framework, i.e., FRL. Our findings reveal that FRL does not guarantee the robustness of the FL system and there are some edges that are particularly vulnerable to poisoning attacks.
Through a theoretical investigation, we prove the existence of these vulnerable edges and establish the lower and the upper bound for identifying them in each layer. 
Building on these insights, we develop a novel local model poisoning attack against FRL, called \underline{V}ulnerable \underline{E}dge \underline{M}anipulation (VEM) attack. This attack comprises two key components: 1) vulnerable edge identification, and 2) optimization-based enhancement. The first component identifies and focuses on the most susceptible parts of the network to ensure a more precise and effective attack. The second component leverages optimization techniques to increase the attack impact.

To execute the VEM attack, we begin by theoretically defining the range of vulnerable edges in each layer. Once these edges are identified, we apply an optimization-based method to maximize the adversarial impact. Specifically, we formalize an optimization function such that the global model changes the most after the attack. Given that updates in FRL are represented as discrete integers, which cannot be directly optimized using standard stochastic gradient descent (SGD) algorithms, we employ the Gumbel-Sinkhorn method \cite{mena2018learning} to transform this discrete problem into a continuous one. This transformation enables us to identify the most effective malicious updates, thereby enhancing the overall attack impact.

We evaluate our attack on eight benchmark datasets and integrate eight Byzantine-robust \textit{aggregation rules} (AGRs) within the FRL framework to test the attack impact of our VEM.
The results demonstrate that our attack achieves a significantly greater attack impact compared to state-of-the-art attacks across all combinations of datasets, model architectures, and AGRs, with an overall attack impact of 53.23\%, making it 3.7 times more effective than other attacks.
Additionally, we developed a custom defense tailored specifically to counter our attack; however, our VEM attack remains effective against this targeted defense. Our key contributions can be summarized as follows:

\begin{itemize}
    \item  We conduct the first systematic analysis of FRL's robustness, uncovering a critical vulnerability within the framework. Our findings reveal that FRL does not guarantee the robustness of the system, and there exists some edges that are particularly vulnerable to poisoning attacks. Additionally, we provide a necessary condition to launch a successful attack in the FRL framework.
    \item Based on the results of the analysis, we design and implement a new attack that targets and effectively manipulates the vulnerable edges. Extensive experiments across different
network architectures and datasets demonstrate that our VEM significantly outperforms SOTA attacks.
\item We release the source code at \url{https://github.com/gongzir1/VEM} to facilitate future studies in this area.
\end{itemize}

\section{Background}
\subsection{FL and Poisoning Attacks on FL}
Federated Learning (FL) \cite{mcmahan2017communication} enables multiple clients to collaboratively train a model without sharing their private training data. This is achieved through a central server that iteratively aggregates local updates computed by the clients.
Consider a cross-device setting, where a set of $\mathbb{N}$ clients jointly train a global model. In each round, the server selects a subset of clients $\mathbb{U} \subseteq \mathbb{N}$ to participate in the current round training and sends the global model $g$ to the selected clients.
Each client then trains its local model $\theta^w$ on its private dataset and sends the model update $\nabla=\theta^w-g$ back to the server. The server aggregates these local updates to produce the global model for the next round of training.

However, the distributed nature makes FL vulnerable to client-side poisoning attacks \cite{biggio2012poisoning,kurakin2016adversarial,munoz2017towards,jagielski2018manipulating,fang2020local,shejwalkar2021manipulating,bagdasaryan2020backdoor,wang2020attack,li20233dfed,zhang2023denial}. Based on the goal of the adversary, they can be categorized into two types: untargeted and targeted attacks. In untargeted poisoning attacks \cite{biggio2012poisoning,kurakin2016adversarial,fang2020local,shejwalkar2021manipulating,zhang2023denial}, the aim is to degrade the overall performance of the global model including its test accuracy, thereby impacting its general reliability and effectiveness. In targeted poisoning attacks \cite{bagdasaryan2020backdoor,wang2020attack,li20233dfed}, where the adversary seeks to lower the model’s accuracy on a specific target class or subset of data while preserving high accuracy on other classes.
Backdoor attacks \cite{bagdasaryan2020backdoor} are a subset of targeted attacks where the attack will only be activated when a backdoor trigger exists.

Based on the adversary's capabilities, poisoning attacks can be categorized as data poisoning attacks and model poisoning attacks. In data poisoning attacks \cite{munoz2017towards,jagielski2018manipulating}, the adversary manipulates the training data to indirectly manipulate the gradients of local update, whereas, in model poisoning attacks (MPA) \cite{jagielski2018manipulating,fang2020local,shejwalkar2021manipulating,bagdasaryan2020backdoor,wang2020attack,li20233dfed}, the adversary directly alters the updated gradients to make the attack more effective.

\subsection{Existing Robust Aggregation Algorithms}
Multiple Byzantine-robust \textit{aggregation rules} (AGRs) are proposed to defend against poisoning attacks \cite{caofltrust,fang2020local,shen2022better,zhang2022fldetector,gong2023agramplifier,karimireddy2021learning,guerraoui2018hidden,sharma2023flair,munoz2019byzantine,yin2018byzantine,xia2019faba,fung2020limitations,nguyen2022flame}. For instance,  Krum \cite{blanchard2017machine} selects one local update that is close to its $U - m - 2$ neighboring update in Euclidean distance as the global model; here, $m$ is an upper bound of the number of malicious clients and $U$ is the number of clients selected for the current training. Multi-Krum~\cite{blanchard2017machine} is based on Krum to select $c$ updates such that $U-c>2m+2$, and then average the selected $c$ updates as the global. Trimmed Mean and Median \cite{yin2018byzantine} employ approach involves coordinate-wise aggregation, where the AGRs separately identify and remove the largest and smallest values in each dimension and aggregate the remaining values. 
Norm-bounding \cite{sun2019can} constrains the $L_2$ norm of all client updates to a fixed threshold before the average aggregation. For a local update $\nabla$ and threshold $T$, if $\|\nabla \|_2 > T$, $\nabla $ will be scaled by $\frac{T}{\|\nabla \|_2}$. This scaling operation effectively reduces the severity of the poisoning level of the malicious model updates. 
More recently, the work in Fang \cite{fang2020local} assumes the server holds a clean validation dataset, and the server can calculate losses and errors on the validation set of the updated model to determine whether the updates originate from malicious or benign clients. We leave the detailed description of other AGRs in Appendix \ref{app:detailed_agr}.

However, these AGRs suffer from two limitations. First, most existing AGRs operate in a continuous space, leaving a large space of potential malicious updates. Second, most of them rely on average or weighted average techniques for aggregation, allowing an individual user to have a disproportionate influence on the global model. 
A recent work, i.e., FRL was proposed to address these issues. 


\subsection{Federated Rank Learning (FRL)}
Before diving into FRL, we first introduce the foundational concept behind it, which is the Lottery Ticket Hypothesis (LTH).

\noindent\textbf{Lottery Tickets Hypothesis and Supermasks.}
Modern neural networks contain an enormous number of parameters, often leading to overparameterization \cite{dauphin2013big,denil2013predicting}.  To solve this issue, 
Frankle and Carbin \cite{frankle2018lottery} offer an intriguing hypothesis i.e., Lottery Tickets Hypothesis (LTH), which states that within a randomly initialized, dense neural network (\textit{supernetwork}), there exists a sparse \textit{subnetwork} such that when trained in isolation, it can achieve the test accuracy as the fully trained original \textit{supernetwork} within the same number of training iterations.
Follow-up work by Ramanujan et al. \cite{ramanujan2020s} proposes the edge-popup (EP) algorithm \cite{ramanujan2020s} to effectively find \textit{subnetworks}. Specifically, the EP algorithm assigns different scores to each edge in the \textit{supernetwork} and uses these scores to determine which edges to retain. The process involves both a forward and a backward pass. During the forward pass, the EP selects the top k\% of edges with the highest scores. In the backward pass, it updates these scores using the straight-through gradient estimator \cite{bengio2013estimating}, enabling gradients to pass through non-differentiable operations. The process is conducted iteratively to update the scores and retain the edges that effectively reduce the training loss. By doing so, it identifies a \textit{subnetwork} that performs as well as the fully trained \textit{supernetwork}.

FRL \cite{mozaffari2023every} expand the LTH hypothesis to the federated learning framework, where clients work collaboratively to select a \textit{subnetwork} that maintains high performance as the original fully trained \textit{supernetwork}.
Specifically, FRL consists of the following three steps: 

\textbf{Server Initialization:}
In the first round, the server chooses a seed to generate random weights $\theta^w$ and scores $\theta^s$ for each edge in the global \textit{supernetwork}; (Note that in the subsequent training, only the scores $\theta^s$ are updated.)
Then, it sorts the scores $\theta^s$ in ascending order to get the corresponding global ranking $R_g$, where each value represents edge ID and the index indicates the edge's \textit{reputation} (with higher indices denoting greater importance). Finally, the server sends the global ranking $R_g$ to the selected $U$ clients.

\textbf{Client Training:} Upon receiving the weights $\theta^w$ and scores $\theta^s$,
the client uses the edge-popup (EP) algorithm to update the score $\theta^s$ for each edge, such that if having an edge $e$ selected for the \textit{subnetwork} reduces the training loss (e.g., cross-entropy loss), the score of edge $e$ increases.
Then, clients sort the updated score and send the updated rankings back to the server for aggregation. 

\textbf{Server Aggregation:} The server receives the local updates $R_u$ for $u \in [1, U]$ and performs majority voting (MV) to get the aggregated global ranking.
Specifically, the server first aligns the \textit{reputation} scores across users by sorting them based on edge IDs.
After alignment, the server performs a dimension-wise summation of these sorted \textit{reputation} to compute the \textit{aggregated reputation} for each edge. Then, the server sorts these \textit{aggregated reputation} scores in ascending order to generate the final ranking of edge IDs.
For instance, if a client's ranking is $R_u=[3, 4, 5, 1, 6, 2]$, where edge $e_3$ is the least important edge (in the first index) and $e_2$ is the most important edge (in the last index), the server assigns \textit{reputation} scores $I_u = [4, 6, 1, 2, 3, 5]$ based on their indices in $R_u$. The server then sums these scores across all users to obtain the \textit{aggregated reputation}, i.e., $W=\sum_{u=1}^{U}I_u$. 
Finally, the server sorts the \textit{aggregated reputation} in ascending order and maps them to their corresponding edge IDs, producing the global ranking $R_g$.
This global ranking can then be transformed into a supermask $\textbf{M}$ (a binary mask of 1s and 0s), which is applied to the random neural network (the \textit{supernetwork}) to generate the final \textit{subnetwork}. 

In summary, the objective of FRL is to find a global ranking $R_g$, such that the resulting \textit{subnetwork} $\theta^w \odot \textbf{M}$ minimizes the average loss of all clients. Formally, we have
\begin{equation}
\begin{alignedat}{2}
& \min_{R_g} F(\theta^w, R_g) = \min_{R_g} \sum_{u=1}^U{\lambda_u \mathcal{L}_u(\theta^w \odot \textbf{M})}, \\
&\text{s.t. } \
\textbf{M}[R_g[i]] = \left\{ \, \begin{IEEEeqnarraybox}[][c]{l?s} \IEEEstrut 0 & if  $i< t$ , \\ 1 &  otherwise, \IEEEstrut \end{IEEEeqnarraybox} \right. \label{eq:example_left_right1} 
\end{alignedat}
\end{equation}
where $U$ denotes the number of clients selected for the current iteration and $\theta^w$ represents the global weight.
$\mathcal{L}_u$ and $\lambda_u$ represent the loss function and the weight for the $u^{th}$ client, respectively, and $t = \left\lfloor n \left(1 - k/100\right) \right\rfloor$ is the index of the selection boundary for the \textit{subnetwork},  with  $k$ indicates the sparsity of the \textit{subnetwork} (i.e., the top $k\%$ of edges are selected for the \textit{subnetwork}) and $n$ is the number of edges in each layer.
Detailed steps can be seen in Algorithm \ref{alg:frl} in Appendix \ref{app:alg}.
\section{Vulnerable Edge Manipulation (VEM)}
\subsection{Threat Model}

\noindent\textbf{Adversary’s Objective.} Since training an FL model is time-consuming and computationally intensive, injecting malicious clients into the training process to destroy the global model can pose a severe threat to the system \cite{shejwalkar2021manipulating}. Therefore, in this paper, we consider the untargeted local model poisoning attack against the FRL framework, such that when the adversary shares the malicious update with the central server, the accuracy of the resulting global model reduces as much as possible.

\textbf{Adversary’s Capabilities.} We assume that the adversary controls up to $m$ out of $U$ clients in each iteration, called malicious clients. Each malicious client has its own clean dataset.
We follow the previous setting \cite{fang2020local,shejwalkar2021manipulating} and assume 
$m/U = 0.2$ unless otherwise stated.

\textbf{Adversary’s knowledge.} We consider the \textit{update-agnostic} setting, where the adversary does not have any information of $U-m$ benign updates. 
However, the adversary will make use of the clean datasets from controlled clients to generate malicious updates.

\subsection{Motivation}
\label{sec:intuition}
\begin{figure}[htbp]
    \centering
    \includegraphics[width=\columnwidth]{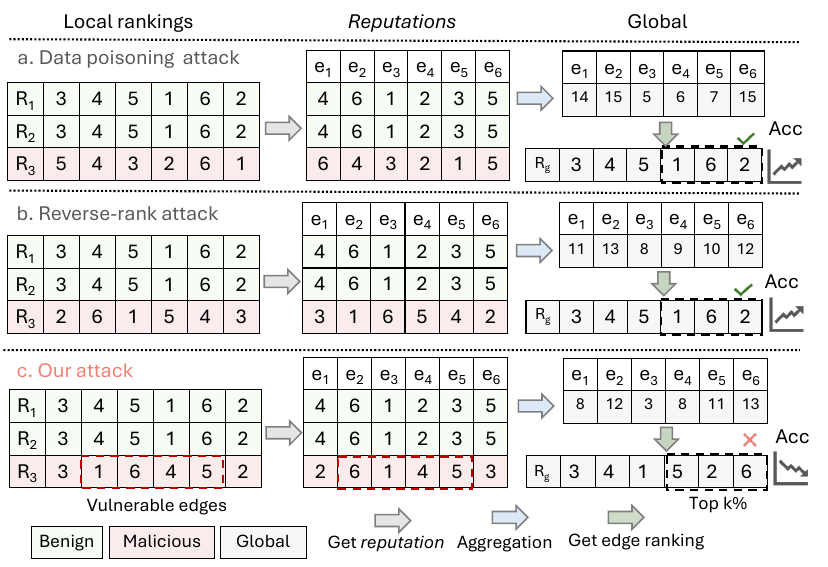}
    \caption{Motivating example of poisoning FRL. a. the `data poisoning attack' where the adversary submits relative random ranking; b. the `reverse-rank' attack where the adversary submits reverse ranking of benign update; c. our attack, which targets and manipulates specific edges. Only our attack leads to changes in the global \textit{subnetwork}.}
    \label{fig:challenges}
\end{figure}
We begin by highlighting the key observations regarding the limitations of existing poisoning attacks on FRL, which motivates the development of our VEM.

\begin{figure}[htbp]
    \centering
    \includegraphics[width=0.95\columnwidth]{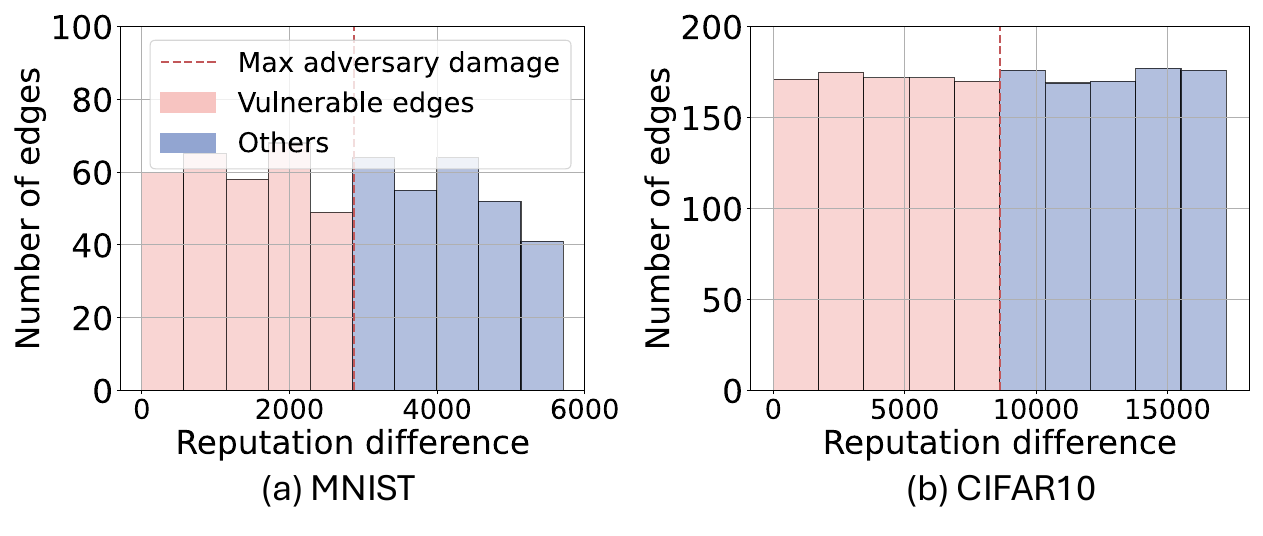}
    \caption{\textit{Reputation} difference between each edge and the selection boundary. }
    \label{fig:difference}
\end{figure}

\begin{figure*}[!t]
    \centering
    \includegraphics[width=\textwidth]{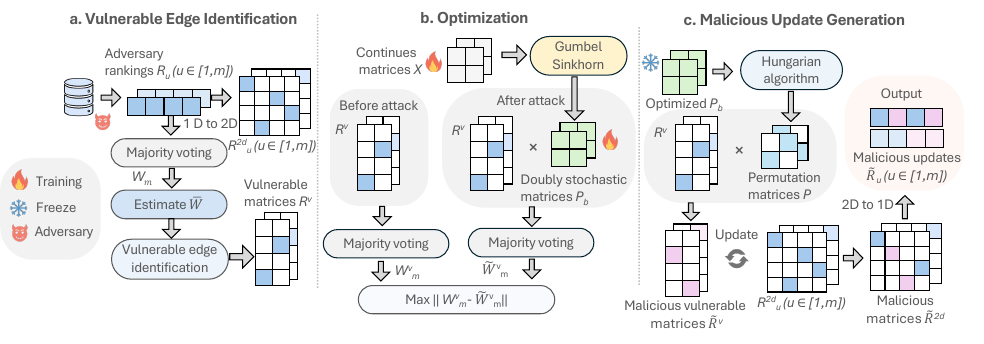}
    \caption{Overview of our VEM attack. The attack unfolds in three stages: a. the adversary identifies vulnerable edges and generates vulnerable matrices for optimization; b. the adversary targets vulnerable edges and finds the optimal doubly stochastic matrices that maximize the \textit{reputation} difference before and after the attack; c. the adversary uses optimized doubly stochastic matrices to produce malicious updates. }
    \label{fig:attack}
\end{figure*}
The update constraints in the FRL system  limit the adversary’s capacity to scale or optimize the malicious updates in the continuous space, making existing model poisoning attacks
(MPA) ineffective against the FRL system. Therefore, to launch poisoning attacks on FRL, we first consider data poisoning attacks where the attacker changes the label of the training dataset to disrupt the consensus of benign updates. For example, in a random label-flip attack \cite{biggio2012poisoning}, the adversary submits a relatively random ranking to the server for aggregation. We illustrate the impact of such attack in Fig. \ref{fig:challenges} (a), where the first two rows represent benign updates (selecting edges [1,~6,~2]), while the third row represents malicious updates. After majority voting, the global model still selects edges [1,~6,~2] for the \textit{subnetwork}, resulting in a potential accuracy increase.
Additionally, in FRL \cite{mozaffari2023every}, the adversary is considered as the `worst-case' untargeted poisoning attack on FRL, where the adversary computes the rankings based on its benign data and uses majority voting to simulate the aggregated ranking. Each malicious client then submits the reverse of this simulated global ranking to the FRL server for aggregation (we refer to this as a reverse-rank attack hereinafter). However, this seemingly strongest approach is still ineffective for poisoning FRL. As shown in Fig. \ref{fig:challenges} (b), even though the malicious user reverses the ranking of every edge, it still results in the selection of edges [1,~6,~2] for the \textit{subnetwork} after majority voting. 
The failure of the above two examples lies in the fact that the ranking-based update imposes an upper bound on the adversary's potential damage, thereby minimizing their influence on the global model selection. Consequently, it becomes difficult for the minority to alter the voting outcomes against the majority.

However, we observe that FRL is not inherently robust, and there are specific edges that are particularly vulnerable to poisoning attacks. 
To verify this argument, we record the \textit{reputation} difference between each edge and the selection boundary $\gamma$, which is calculated as $\Delta_j = |w_j - \gamma|$, and $w_j$ is the \textit{aggregated reputation} for edge $e_j$ (for $j \in [1, n]$). 
This metric indicates the minimal modification needed for edge $e_j$ to cross the selection boundary, thereby reflecting its vulnerability. A lower value of $\Delta_j$ signifies a higher vulnerability.
As illustrated in Fig. \ref{fig:difference}, there is a significant variation in $\Delta_j$ (for $j \in [1, n]$), ranging from 0 to 5616 for the MNIST dataset and from 0 to 17208 for the CIFAR10 dataset.  (Details of the experiment setting can be seen in Appendix \ref{app:fig3}.) Additionally, we use the red dash line to indicate the max adversary damage (e.g., 20\% malicious rate). Edges to the left of this line are considered vulnerable, as the required reputation change is smaller than the maximum adversary damage.

Based on the above observation, we update malicious ranking by manipulating only the vulnerable edges. As a result, the global model selects edges [5,~2,~6] for the \textit{subnetwork}, leading to a potential accuracy drop, as shown in Fig. \ref{fig:challenges} (c).

\subsection{VEM Overview}
Our VEM aims to manipulate malicious updates such that the aggregated global model deviates significantly from their original ones at each training round, thus negatively affecting the global model performance. 
To effectively achieve this goal and make the FRL framework tackleable, we first provide a formalization of FRL in Section \ref{sec:formalize}. Specifically, we expand the 1D ranking into a 2D matrix to reduce the non-differentiable operations involved in the original algorithm.

Based on the formalization, our attack unfolds in three main stages: vulnerable edge identification, optimization, and malicious update generation (shown in Fig. \ref{fig:attack}).
In the first stage, the adversary aims to identify vulnerable edges within each layer. To achieve this, it uses the rankings generated from its own datasets and simulates the majority voting process locally. After getting the aggregated ranking, it incorporates the global ranking from the previous round to estimate vulnerable edges. Once identify the vulnerable edges, the adversary extracts the ranking of those vulnerable edges to form vulnerable matrices (details see Section \ref{sec:find_vulnerable}).
In the optimization stage, the adversary aims to target those vulnerable edges and form the optimization function such that the global models \textit{reputation} of those vulnerable edges deviate significantly from their original values. To solve the optimization function, we use the Gumbel-Sinkhorn method to convert a discrete problem into a continuous problem (details see Section \ref{sec:optimization}). After the optimization process, we use the optimized parameter to generate malicious vulnerable matrices, which are then used to produce malicious updates (details see Section \ref{sec:generation}). 

\subsection{Formalization}
\label{sec:formalize}
In this section, we formalize FRL in a practical manner. Specifically, we outline the formalization in two key steps.

\noindent{\textbf{1D to 2D:}}
Since the ranking $R$ stores two-dimensional information (i.e., the value indicates the edge ID, and the index indicates its \textit{reputation}), in the formalize, we first expand $R$ into a two-dimensional matrix $R^{2d}$, to remove the non-differentiable operation in the original FRL. 
This expansion is based on the fact that there is a natural one-to-one correspondence between permutations and permutation matrices, and the permutation in one form can be converted to another form \cite{zhang2018improved}. Formally, we have the following definition.

\begin{figure}[htbp]
    \centering  \includegraphics[width=0.95\columnwidth]{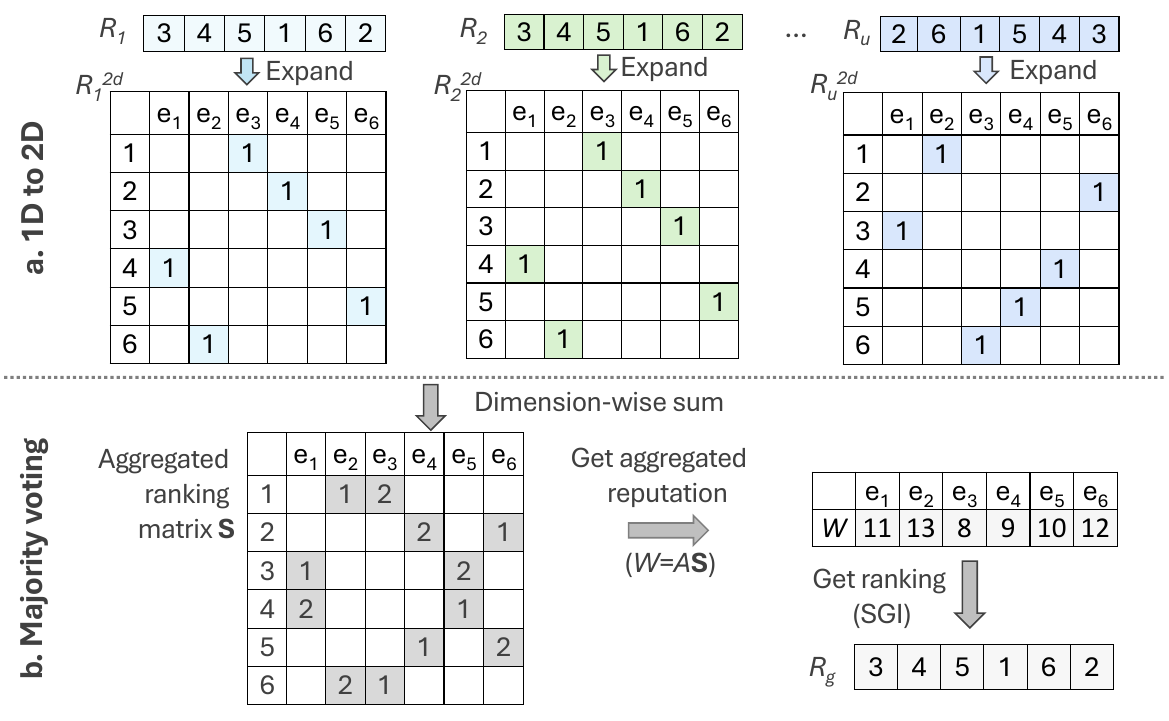}
    \caption{Formalization of FRL.}
    \label{fig:2d}
\end{figure}

\noindent\textbf{Definition 1} (Permutation Matrix).
Given a permutation $R$ of \(n\) elements, the corresponding permutation matrix $R^{2d}$ is an \(n \times n\) matrix defined as follows:
\begin{equation}
R^{2d}[i, j] = \begin{cases} 
1 & \text{if } R[i] = j, \\
0 & \text{otherwise}.
\end{cases}
\end{equation}
In $R^{2d}$, the column $j$ indicates the edge ID, and the row $i$ indicates the \textit{reputation}. 
For instance, as shown in Fig. \ref{fig:2d} (a), given $R_1=[3, 4, 5, 1, 6, 2]$, meaning that edge $e_3$ is the least important edge (in the first index) and $e_2$ is the most important edge (in the last index). After the dimension expansion, 
$R^{2d}_1[1, 3]=1$ meaning that edge $e_3$ is in the first index in $R_1$, $R_1^{2d}[6, 2]=1$ meaning that $e_2$ is in the sixth index (for better visibility, we leave all 0s empty).

\noindent{\textbf{Majority Voting:}} Based on the dimension expansion, we further formalize the majority voting process.
Specifically, after receiving $R^{2d}_u$ for $u \in [1,U]$, the server first calculates a dimension-wise sum, i.e., $\mathbf{S} = \sum_{u=1}^{U} R^{2d}_u$ to get the aggregated ranking matrix $\mathbf{S}$.
In $\mathbf{S}$, each column $j$ indicates an edge ID, and each row $i$ indicates the \textit{reputation} score, and the value at each position indicates the number of votes.
As shown in Fig. \ref{fig:2d} (b), $\mathbf{S}[1, 3]=2$, which means two clients agree that edge $e_3$ has a least \textit{reputation}. $\mathbf{S}[2, 4]=2$ indicate that two clients think $e_4$ has the second least reputation.
Then, for each edge $e_j$,  we can get the \textit{aggregated reputation} by
\begin{equation}
\begin{alignedat}{2}
\label{eq:weight}
w_j&=\sum_{i=1}^{n} a_i s_{i,j} =A  s_j,
\end{alignedat}
\end{equation}
where $s_j$ is the $j^{th}$ column in $\mathbf{S}$, $A=[a_1, a_2, \dots, a_{n}]$ is a asceniding vector indicating the \textit{reputation} for each index. Following the design of original FRL \cite{mozaffari2023every}, we use $a_i=i$, resulting in $A=[1, 2, \dots, n]$. 
Lastly, the server performs the sort-get-index  ($\mathrm{SGI}$) operation to get the ranking of global model $R_g$ and selects top $k\%$ edges for the \textit{subnetwork}.
Therefore, the majority voting function can be formulated as follows:
\begin{equation}
\begin{alignedat}{3}
\label{eq:wmv}
   R_g =\mathrm{SGI}(A \sum_{u=1}^{U} R^{2d}_u) 
    =\mathrm{SGI} (A  \mathbf{S}) 
    =\mathrm{SGI} (w_1,w_2, \dots, w_n),
\end{alignedat}
\end{equation}
where $w_j$ denote the \textit{aggregated reputation} score for edge $e_j$ ($j \in [1, n]$).
In the following section, we denote $W = [w_1, w_2, \dots, w_n]$ as the \textit{aggregated reputation} for each edge. 
$\mathbb{W}^{\text{in}} = \{w_j | w_j \in W, \text{where} \ e_j \in \mathbb{E}^{\text{in}} \}$ represents the \textit{aggregated reputation} for top k\% edges and $\mathbb{W}^{\text{out}} = \{w_j | w_j \in W, \text{where} \ e_j \notin \mathbb{E}^{\text{in}} \}$ represents the \textit{aggregated reputation} for other edges, and $\mathbb{E}^{\text{in}}$ is the set of top k\% edges selected for the \textit{subnetwork}.




\subsection{Vulnerable Edges Identification}
\label{sec:find_vulnerable}
In this section, we provide a theoretical analysis of the range of vulnerable edges.
Intuitively, if the \textit{reputation} difference $\Delta_j$ between edge $e_j$ and the selection boundary is smaller than the maximum damage the adversary can cause in one round, we call the edge $e_j$ a vulnerable edge. 
Formally, we have the following theorem. \\

\begin{figure}[htbp]
    \centering
    \includegraphics[width=0.93\columnwidth]{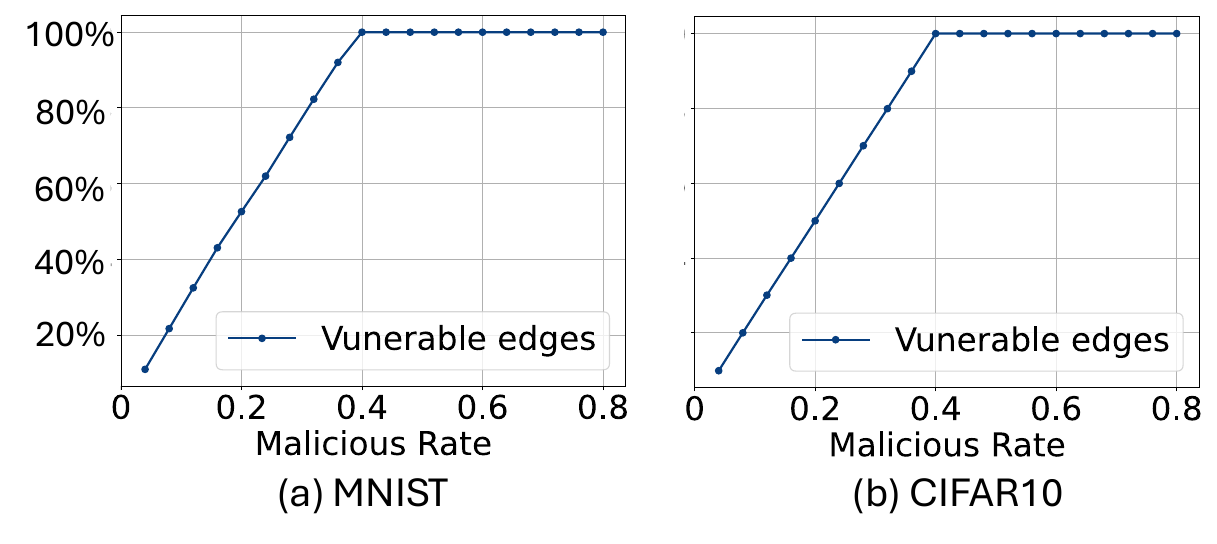}
    \caption{Percentage of vulnerable edges among all edges with different malicious rates.}
    \label{fig:implication}
\end{figure}
\textbf{Theorem 1.}
\textit{Give the aggregated reputation of $U-m$ benign users, i.e., $\overline{W}=[\overline{w}_1, \overline{w}_2, \dots,\overline{w}_n]$,
the \textit{reputation} of a vulnerable edge $e_v$ is bounded by}
\begin{equation}
\label{eq:theorem}
   \overline{w}_{\text{max}} - m(a_n-a_1) <\overline{w}_v<\overline{w}_{\text{min}}+m(a_n-a_1),
\end{equation}
\textit{where} $\overline{w}_{\text{min}}=\mathrm{\text{min}}(\overline{\mathbb{W}}^{\text{in}})$ \textit{and} $\overline{w}_{\text{max}}=\mathrm{\text{max}}(\overline{\mathbb{W}}^{\text{out}})$.

Below, we discuss the implication of the theorem, and we leave its proof to Appendix \ref{proof:theorm1}.\\
\textbf{Implications.} 
Given that the user input is discrete integers ranging from 1 to $n$, the one-round adversary's damage is limited by a maximum budget i.e., $m(a_n-a_1)$ for $m$ malicious users. Therefore to launch a successful attack for each round (i.e., move edges across the selection boundary), a necessary condition is that the needed \textit{reputation} change, i.e., $\overline{w}_{\text{min}}-\overline{w}_v$ or $\overline{w}_{\text{max}}-\overline{w}_v$ should not exceed the one-round adversary's damage budget. Therefore, the edges that the adversary can alter (i.e., vulnerable edges) are bounded by Eq.~(\ref{eq:theorem}).
Moreover, as $m$ increases, the range of vulnerable edges expands. 
As shown in Fig. \ref{fig:implication}, the x-axis denotes the malicious rate and the y-axis indicates the percentage of vulnerable edges calculated based on Theorem 1. As the malicious rate rises, so does the percentage of vulnerable edges. When the malicious rate reaches 0.4, all edges can be considered vulnerable. Even with an extreme scenario where just 1 malicious client (4\% malicious rate), there are still 11\% and 10\% vulnerable edges on the MNIST and CIFAR10 datasets, respectively.

Based on the given Theorem 1, we define $\overline{w}_{\epsilon_1} = \overline{w}_{\text{max}} - m(a_n-a_1)$ as the lower bound for vulnerable edges and $\overline{w}_{\epsilon_2} =  \overline{w}_{\text{min}}+m(a_n-a_1)$ as the upper bound.
Then we define the ranking of vulnerable edges as follows:
\begin{equation}
R^v[i,j]= \begin{cases} 
R^{2d}[i,j], & \text{if }  \overline{w}_{\epsilon_1} < \overline{w}_j <  \overline{w}_{\epsilon_2}, \\
0, & \text{otherwise}.
\end{cases}
\end{equation}
We call $R^v$ vulnerable matrices in the following section.
Additionally, to enhance the stealthiness of VEM, we introduce a hyper-parameter $\zeta$ to adjust the range of identified vulnerable edges, where $\zeta \in (0,1]$. Therefore we have

\begin{equation}
\label{eq:with-zeta}
R^v[i,j] = 
\begin{cases} 
R^{2d}[i,j], & \text{if } \overline{w}_{\epsilon_1} + (1 - \zeta) (\overline{w}_{\epsilon_2} - \overline{w}_{\epsilon_1})/2 < \overline{w}_j \\
& < \overline{w}_{\epsilon_2} - (1 - \zeta) (\overline{w}_{\epsilon_2} - \overline{w}_{\epsilon_1})/2, \\[10pt]
0, & \text{otherwise}.
\end{cases}
\end{equation}
A higher $\zeta$ results in a more powerful attack, whereas a lower $\zeta$ results in a more covert attack. Unless specified otherwise, we assume $\zeta = 1$, indicating that all identified vulnerable edges are manipulated.

\noindent{\textbf{Benign Updates Estimation.}}
\label{sec: estimation}
As illustrated in Theorem 1, identifying vulnerable edges requires the knowledge of \textit{aggregated reputation} $\overline{W}$ of $U-m$ benign users. However, we assume the adversary only has $m$ updates generated using its own dataset. Therefore, in the following section, we discuss two ways for estimating $\overline{W}$, namely alternative estimation and historical estimation.

For the alternative estimation, we use the aggregation of $m$ updates to get an estimated value of the aggregated \textit{reputation} of other $U-m$ users,
such that we have
\begin{equation}
\label{eq:estimiated_threshold_alternative}
\overline{W} \approx (U/m-1)  \cdot \left( A \sum_{u=1}^{m} {R}^{2d}_u
\right), 
\end{equation}
where $R_{u}^{2d}$ (for $u \in [1, m]$) are generated using adversary's own dataset. 
This method is widely adopted in previous MPA attacks \cite{fang2020local,shejwalkar2021manipulating,zhang2023denial} of the FL system. 
For the historical estimation, we use the global ranking $R^{2d}_g(t-1)$ and malicious update  $\widetilde{R}_{u}^{2d}(t-1)$ in $t-1$ round to estimate $\overline{W}(t)$ \footnote{Note that we omit $t$ in other notion for better visibility and unless specified other, we denote the current round.}. Formally, we have
\begin{equation}
\begin{alignedat}{2}
\label{eq:estimiated_threshold_history}
\overline{W}(t) &\approx \overline{W}(t-1) = W(t-1)- \widetilde{W}(t-1), \\
&\approx U \cdot \left( A R_g^{2d}(t-1) \right)- \left(A \sum_{u=1}^m \widetilde{R}_u^{2d}(t-1)  \right).
\end{alignedat}
 \end{equation}
\noindent\textbf{Use Case for Two Estimation Methods.}
In the IID case, both methods can achieve the desired estimation performance. In the non-IID case, the historical estimation method offers more reliable performance, as it leverages the global ranking from the previous round for estimation. 
We defer detailed discussion of these two estimation methods in Section \ref{sec:eva_estimation}.

\subsection{Optimization}
\label{sec:optimization}
After identifying the vulnerable edges, we aim to target those edges and manipulate their ranking, thus effectively affecting the global model performance. 
Therefore, we have the following optimization function such that the global models \textit{reputation} of those vulnerable edges deviate significantly from their original values for each training round. Formally, we have
\begin{equation}
\label{eq:objective}
\begin{alignedat}{2}
&\arg\max_{\widetilde{R}_1^v,\widetilde{R}_2^v,\dots,\widetilde{R}_m^v} \left\| W^v- \widetilde{W}^v \right\|, \\  \text{s.t.} \
&W^v=A \sum_{u=1}^{U} R_u^v, \\
&\widetilde{W}^v=A (\sum_{u=1}^{m} \widetilde{R}_u^v+\sum_{u=m+1}^{U} R_u^v),
\end{alignedat}
\end{equation}
where $R^v_u$ (for ${u \in [1, U]}$) denotes the ranking of vulnerable edges for benign updates, while $\widetilde{R}^v_u$ (for ${u \in [1, m]}$) denotes the ranking of vulnerable edges for malicious updates. 
$W^v$ represents the \textit{aggregated reputation} of vulnerable edges before an attack, and $\widetilde{W}^v$ represents it after the attack.
$|| \cdot ||$ denote the Euclidean norm.

However, Eq. (\ref{eq:objective}) can not be solved using the SGD-based method since $\widetilde{R}^v_u$ is not continuous. Therefore, instead of optimizing $\widetilde{R}^v_u$ directly, we solve Eq. (\ref{eq:objective}) by applying a permutation operation to the benign updates to obtain the malicious updates, such that we have $\widetilde{R}^v_u= R^v_u  P_u$ (for $u \in [1, m]$), where $P_u$ is the permutation matrix.
Additionally, since $\sum_{u=m+1}^{U} R_u^v$ remains unchanged before and after the attack, we omit this term. 
Therefore, Eq. (\ref{eq:objective}) can be re-formulated as 
\begin{equation}
\label{eq:object-3}
\begin{alignedat}{5}
& \arg\max_{P_1,P_2,\dots,P_m} \left\| W_m^v- \widetilde{W}_m^v \right\|, \\
 \text{s.t.} \
&W_m^v=A \sum_{u=1}^{m} R_u^v, \\
&\widetilde{W}_m^v=A \sum_{u=1}^{m} (R_u^v P_u),
\end{alignedat}
\end{equation}
where $W_m^v$ is the \textit{aggregated reputation} of $m$ updates to vulnerable edges before the attack and $\widetilde{W}_m^v$ is that after the attack.\\
\noindent{\textbf{Solve Optimization Function.}}
Since the permutation matrix is discrete and cannot be optimized directly, a practical way to solve Eq. (\ref{eq:object-3}) is to use a continuous matrix $X$ to approximate the discrete matrix. In the training process, optimizing $X$ will simultaneously update the permutation matrix $P$.
Therefore, we have the following definition:\\
\textbf{Definition 2} (Matching Operation) \cite{kuhn1955hungarian}. Given a matrix $X$ drawn from a continuous distribution, we define the matching operator $\mathcal{M}(X)$ as follows:
\begin{equation}
\label{eq:matching}
\mathcal{M}(X) = \arg \max_{P \in \mathbb{P}} \langle P, X \rangle_F,
\end{equation}
where $\mathbb{P}$ denoting the set of permutation matrices and $\langle P, X \rangle_F = \text{trace}(A^\top B)$ is the (Frobenius) inner product of matrices.

However, the matching operation $\mathcal{M}(X)$ is still not differentiable, so we seek a continuous method to approximate this discrete operation. To achieve this, we employ the Gumbel-Sinkhorn approach \cite{mena2018learning} to approximate this matching operation.
Specifically, we first use the Sinkhorn operator \cite{adams2011ranking} to iteratively normalize the rows and columns, resulting in a doubly stochastic matrix. Formally, we have
\begin{equation}
\label{eq:Sinkhorn}
\begin{alignedat}{3}
S_0(X)& = \exp(X), \\
S_l(X)& = T_c(T_r(S_{l-1}(X))), \\
S(X)& = \lim_{l \to \infty} S_l(X),
\end{alignedat}
\end{equation}
where $T_r(X) = X \oslash (X 1_V 1_V^\top)$
and $T_c(X) = X \oslash (1_V 1_V^\top X)$ as the row and column-wise normalization operators, respectively. 
 \( 1_V \) is a column vector of ones of length \( \ (\epsilon_2 - \epsilon_1)\), \( 1_V^\top \) is its transpose (a row vector of ones), and \( \oslash \) denotes element-wise division.
Then, we can use the following continuous method to approach Eq. (\ref{eq:matching}):
\begin{equation}
S(X/\tau) = \arg \max_{P^b \in \mathbb{B}} \langle P^b, X \rangle_F + \tau h(P^b),
\end{equation}
where $\mathbb{B}$ is the set of doubly-stochastic matrices, $\tau$ is a positive temperature parameter, and 
$h(P^b)$ is the entropy of matrix $P^b$, and is calculated as $h(P^b) = - \sum_{i,j} P^b[i,j] \log (P^b[i,j]))$.
And \(\mathcal{M}(X)\) can be obtained as the limit of \(S(X/\tau)\), meaning that one can approximate \(\mathcal{M}(X) \approx S(X/\tau)\) with a small \(\tau\), formally we have
\begin{equation}
\label{eq:lim}
\mathcal{M}(X) = \lim_{\tau \to 0^+} S(X/\tau).
\end{equation}
For detailed proof, please refer to \cite{mena2018learning}.

Note that the analytical solution of Eq.~(\ref{eq:lim}) cannot be obtained, 
as it involves a limit on the iterations $l$ in Eq. (\ref{eq:Sinkhorn}) and the temperature $\tau$ in Eq. (\ref{eq:lim}). 
Instead, after the optimization process, we obtain the optimized doubly stochastic matrices $P^b$, which approximate the analytical solution and thus maximize the \textit{reputation} difference. 

\subsection{Malicious Update Generation}
\label{sec:generation} 
After getting the optimized doubly stochastic matrices $P^b$, we use them to generate the final malicious update. 
First, we apply the Hungarian algorithm \cite{mills2007dynamic} to derive the permutation matrices $P$ from the doubly stochastic matrices $P^b$. Specifically, we treat $P^b$ obtained from Eq. (\ref{eq:lim}) as the cost matrices and use the Hungarian algorithm to find $P$ that maximizes the cost score, i.e., $\text{trace}(P^\top P^b)$.
Once we have the permutation matrices $P$, we use them to permute the original vulnerable matrices $R^v$ to obtain the malicious vulnerable matrices $\widetilde{R}^v$. Finally, we use $\widetilde{R}^v$ to update the ranking of vulnerable edges in $R^{2d}_u$ for $u \in [1, m]$.

Overall, we summarize our attack process in Algorithm \ref{alg:attack}.
In the first stage, the adversary aims to simulate the majority voting process locally to get the estimated vulnerable edges. Specifically, the adversary first calculates local rankings $R_u$ (for $u \in [1, m])$ using its own dataset and expands them to get 2D ranking $R^{2d}_u$ (for $u \in [1, m])$ (lines 2 - 5).
The adversary then employs majority voting to compute the \textit{aggregated reputation} $W_m$ of $m$ updates (line 6). Subsequently, it estimates the \textit{aggregated reputation} $\overline{W}$ of the other $U - m$ benign clients (line 5) using global ranking $R_g(t-1)$ from the previous round (line 7), and uses it to calculate the range of vulnerable edge ($\epsilon_1, \epsilon_2$) (line 8). After determining the vulnerable edge, it extracts vulnerable matrices $R^v$ from $R^{2d}_u$ (for $u \in [1,m])$ (line 9).
In the optimization stage, the adversary first generates $m$ continuous matrices $X$ with the size of $(\epsilon_2 -\epsilon_1) \times (\epsilon_2 -\epsilon_1)$ (line 11) and employs the Gumbel-Sinkhorn algorithm to map them to doubly stochastic matrices $P^b$, which are used to generate malicious ranking $\widetilde{R}^v$ for vulnerable edges. Then the adversary uses the optimizer to get the optimal $X$ that solves Eq. (\ref{eq:object-3}), which will simultaneously get the optimal $P^b$ (lines 12 - 15).
After getting the optimal doubly stochastic matrices $P^b$, the adversary applies the Hungarian algorithm to convert $P^b$ into the permutation matrices $P$ (line 17) and then uses the $P$ to permute $R^v$, resulting in the malicious ranking of the vulnerable edges $\widetilde{R}^v$ (line 18). Finally, the adversary updates the malicious rankings using $\widetilde{R}^v$ (line 19) to get the malicious matrices $\widetilde{R}^{2d}$ and sends the dimension-reduced malicious ranking $\widetilde{R}_u$ (for $u \in [1, m]$) to the server for aggregation. 
\begin{algorithm}[htbp]
\caption{Vulnerable Edge Manipulation (VEM) attack.}
\label{alg:attack}
\begin{algorithmic}[1]
\renewcommand{\algorithmicrequire}{\textbf{Input:}}
\renewcommand{\algorithmicensure}{\textbf{Output:}}

\Require $D_u$ malicious training datasets,
learning rate for edge pop-up $\eta_\text{ep}$,
$m$ benign rankings $R^{2d}$, $U$ number of clients selected for current round, $m$ number of malicious clients, $k$ \textit{subnetwork} sparsity, number of epoch $E$, $L$ iterations for Gumbel-Sinkhorn training, $\tau$ temperature factor for Gumbel-Sinkhorn,
learning rate for attack $\eta_\text{attack}$.

\Ensure $\widetilde{R}_1,\widetilde{R}_2,\dots,\widetilde{R}_m$
\State {// Vulnerable edge identification}
\For{each $u \in [1, m]$}
    \State $R_u \gets \textsc{Local\_Training}(D_u, \theta^w, \theta^s)$
    \State $R^{2d}_u \gets \Call{1d\_to\_2d}{R_u}$
\EndFor
\State $W_m \gets \textsc{Majority\_Voting}(R^{2d}_1, \dots, R^{2d}_m, A)$
\State $\overline{W} \gets \Call{Estimation} {W_m, R_g(t-1)}$
\State $\epsilon_1, \epsilon_2 \gets \Call{get\_Vulnerable\_edge}{\overline{W}, A, m, k\%}$
\State $R^v \gets \Call{get\_Vulnerable\_matrices}{R^{2d}_1, \dots, R^{2d}_m, \epsilon_1, \epsilon_2}$ 
\State {// Optimization:}
\State $X \gets \text{m continues matrices of size }  (\epsilon_2 - \epsilon_1) \times (\epsilon_2 - \epsilon_1)$
\For{$e \in [1,E]$}
    \State $P^b \gets \Call{Gumbel\_Sinkhorn}{X, L, \tau}$
    \State $X=X- \eta_\text{attack} \nabla \ell(R^v,X,A)$
\EndFor
\State {// Malicious update generation}
\State $P \gets \Call{Hungarian\_algorithm}{P^b}$
\State $\widetilde{R}^v=R^v P$
\State $ \widetilde{R}^{2d}\gets \Call{update}{R^{2d}_1, \dots, R^{2d}_m, \widetilde{R}^v,}$
\For{each $u \in [1, m]$}
    \State  $\widetilde{R}_u^{2d} \gets \widetilde{R}^{2d}[u]$
    \State $\widetilde{R}_u \gets \Call{2d\_to\_1d}{\widetilde{R}^{2d}_u}$
\EndFor
    
\end{algorithmic}
\end{algorithm}
\vspace{-2mm}

\section{Empirical Evaluation}

\subsection{Experimental setup}
\label{sec: set_up}
\noindent\textbf{Datasets.}
For our main experiment, we use four benchmark datasets widely used in prior works to evaluate our attacks, including MNIST \cite{lecun1998mnist}, CIFAR10 \cite{krizhevsky2009learning}, FashionMNIST \cite{xiao2017fashion}, and EMNIST \cite{cohen2017emnist}.  Details are deferred to Appendix \ref{sec:detailed_dataset}.
We use an independently and identically distributed dataset (IID) because poisoning FL with IID data is the hardest \cite{fang2020local,shejwalkar2021manipulating}. 
We also conduct experiments on CIFAR100 \cite{krizhevsky2009learning}, Location30 \cite{yang2016participatory}, Purchase100 \cite{shokri2017membership} and Texas100 \cite{shokri2017membership}, with the results presented in Appendix \ref{app: addtional_exp}.

\noindent\textbf{FL Setting.}
By default, we consider a cross-device FL, where $1000$ clients collaborate to train a global model. In each training round, the server randomly selects 25 clients to participate. We run the experiment for 500 global rounds.

\noindent\textbf{Attack Setting.}
For the attack optimization, we use the Adam optimizer with a learning rate of 0.1 for 50 epochs. Unless stated otherwise, we assume that 20\% of the clients are malicious.  

\noindent\textbf{Evaluation Metric.} We follow previous works \cite{fang2020local,shejwalkar2021manipulating} and conduct our evaluations by simulating the attack on our local device. We follow \cite{zhang2023denial} to use the attack impact to measure the attack effect. Specifically, we use $\text{acc}_{\text{benign}}$ to denote the accuracy of the best global model in the benign setting, and $\text{acc}_{\text{drop}}$ represent the accuracy degradation caused by the attack. The attack impact is then defined as $\phi = \text{acc}_{\text{drop}}/\text{acc}_{\text{benign}} \times 100\%$. A higher $\phi$ indicates a more successful attack.

Additionally, we define $\rho = (|\mathbb{E}^{\text{in}}|-|\mathbb{E}^{\text{in}} \cap \widetilde{\mathbb{E}}^{\text{in}}|)/|\mathbb{E}^{\text{in}}|$ as the percentage of edges that cross the selection boundary caused by the attack, where $\mathbb{E}^{\text{in}}=\{e_j \mid w_j \geq \gamma\}$ be the top $k\%$ edges selected by the \textit{subnetwork} before the attack and
$\widetilde{\mathbb{E}}^{\text{in}}=\{e_j \mid \widetilde{w}_j \geq \widetilde{\gamma}\}$ be the top $k\%$ edges selected by the \textit{subnetwork} after the attack. 
We refer to this metric as `edge cross rate' ($\rho$) hereinafter. A higher $\rho$ implies a more successful attack.

\noindent\textbf{{Poisoning Attacks for Comparison.}}
\label{sec: attacks}
We compare our VEM with six attacks, including Label-flip attack \cite{biggio2012poisoning}, Noise attack, Grad-ascent attack \cite{kurakin2016adversarial}, Min-max attack \cite{shejwalkar2021manipulating}, Min-sum attack \cite{shejwalkar2021manipulating}, Rank-reverse attack \cite{mozaffari2023every}, and we leave the detailed description for each attack in Appendix \ref{sec:detailed_attack}.
For Min-max, Min-sum and Fang attacks, we relax the adversary assumption and assume the adversary has full knowledge of benign updates (as required by the attacks). 
For the Noise attack, Grad-ascent attack, Min-max attack, and Min-sum attack, which are optimized in continuous space, we adapt these attacks to FRL by implementing the attack strategy on the importance scores $\theta^s$ to achieve a similar attack effect. Note that this represents the most effective implementation feasible within the scope of our study.

\subsection{Byzantine-robust Aggregation Rules}
\label{sec:agrs}

\begin{table}[]
\centering
 \caption{Different types of AGRs and their applicability to integrate with FRL.  \protect\blackcircle{} indicates the AGR is directly applicable for integration, \protect\halfblackwhitecircle{} indicates the AGR needs modification for integration, and \protect\whitecircle{} indicates not applicable for integration.}
 \label{tab:agrs}
 \resizebox{0.85\columnwidth}{!}{
\begin{tabular}{c|c|c}
\toprule
\textbf{Category} & \textbf{AGRs} & \textbf{Applicability} \\ \midrule
\multirow{5}{*}{Distance-based} & Krum \cite{blanchard2017machine} & \blackcircle \\ 
                          & Multi-Krum \cite{blanchard2017machine} & \blackcircle \\ 
                          & AFA \cite{munoz2019byzantine} & \blackcircle \\ 
                          & FABA \cite{xia2019faba} & \blackcircle \\ 
                          & DnC \cite{shejwalkar2021manipulating} & \blackcircle \\ \midrule
\multirow{3}{*}{Dimension-wise} & Median \cite{yin2018byzantine} & \whitecircle \\ 
                          & Bulyan \cite{guerraoui2018hidden} & \whitecircle \\ 
                          & TrMean \cite{yin2018byzantine} & \whitecircle \\ \midrule
\multirow{2}{*}{Norm-bounded} & Norm bound \cite{sun2019can} & \whitecircle \\ 
                      & Centered Clip \cite{karimireddy2021learning} & \whitecircle \\ \midrule
\multirow{3}{*}{Weighted-aggregation} & FLTrust \cite{caofltrust} & \halfblackwhitecircle \\ 
                          & FLAIR \cite{sharma2023flair} & \halfblackwhitecircle \\ 
                          & FoolsGold \cite{fung2020limitations} & \halfblackwhitecircle \\ \midrule
\multirow{2}{*}{Validation-based} & Fang \cite{fang2020local} & \halfblackwhitecircle \\  
                            & FLDetector \cite{zhang2022fldetector} & \whitecircle \\ \bottomrule
\end{tabular}}
\end{table}

Following we consider our attack against the Byzantine-robust Aggregation Rules (AGRs) integrated FRL.
Table \ref{tab:agrs} illustrates the five types of AGRs and their applicability to integrated with FRL.  
For Distance-based and Validation-based AGRs, we directly apply those to the user updates (rankings).
For the weighted-aggregation-based AGRs, we first calculate the trust score $s^{agr}_u$ for $u \in [1, U]$ using each AGR strategy and then use the following weighted majority voting to get the global ranking:
\begin{equation}
\begin{alignedat}{3}
\label{eq:weighted_agr}
   R_g &=\mathrm{SGI}(A  \sum_{u=1}^{U} (s^{agr}_u  R^{2d}_u)). \\
\end{alignedat}
\end{equation}
However, the Norm-bounded AGRs restrict the norm of local updates to a fixed threshold before aggregation, while dimension-wise AGRs identify outliers in each dimension. Therefore these approaches are unsuitable to integrate with FRL.
Additionally, FLDetector is based on $\nabla_t=\nabla_{t-1}+ H_t  (g_t -g_{t-1})$, where $\nabla_t$ indicate the user update in iteration $t$, and $g_t$ indicate global model in iteration $t$, and $H_t = \int_0^1 H \left( g_{t-1} + x (g_t - g_{t-1}) \right) \, dx$ is an integrated Hessian for client in iteration $t$. Due to the differing aggregation rules from FRL, we do not consider FLDetector in our experiments.
Regarding Krum-integrated FRL, the convergence for the clean model is not stable due to the large portion of discard. Therefore, we do not include that in the results of our experiments. For detailed descriptions of each AGR, please refer to Appendix \ref{app:detailed_agr}. Beyond AGRs, we also discuss our attack in the context of other defense methods, including augmented defenses and certified defenses, details see Appendix \ref{app:other_defense}.

\section{Experiment Result}
\label{sec:exp}
\subsection{Comparison with the State-of-the-art Attacks}
\label{sec: main_result}
\begin{table*}[!h]
  \centering
  \caption{Comparing state-of-the-art poisoning attacks and our VEM under FRL and AGR integrated FRL. The `No attack’ column reports the clean model accuracy $\text{acc}_{\text{benign}}$ (\%); other columns 
  report the attack impact $\phi$ (\%). For each row, we bold the strongest attack.}
  \label{tab:attack-impact}
  \resizebox{0.85\textwidth}{!}{%
  \begin{tabular}{c|c|c|ccccccc}
    \toprule
    Dataset & AGRs & No attack & Label-flip  & Noise  & Grad-ascent & Min-max  & Min-sum  & Reverse-rank & VEM \\ \midrule
    \multirow{8}{*}{\centering \begin{tabular}{c} MNIST\\(Conv2)\end{tabular}}
    & FRL           & 98.63           &   3.19                  & 3.68                & 1.30                    &  0.45               &   0.79       & 0.64                & \textbf{55.80}     \\
    & Multi-krum    & 95.42           & 2.24                    & 0.00               & 0.36                    & 5.20                & 3.14         & 7.76                & \textbf{48.17}     \\
    & AFA           & 95.38           & 0.00                    & 1.94               & 0.74                    & 9.05                & 6.07         & 5.19                & \textbf{58.88}     \\
    & Fang   & 96.88           & 3.86                    & 1.73               & 1.11                    & 7.01                & 6.95         & 8.53                &\textbf{ 40.70  }   \\
   &FABA &97.39 &0.18 &0.02 &1.43 &5.84 &7.71 &4.29 &\textbf{50.19}\\
   &DnC &98.98 &1.00 &0.01 &0.93 &0.87 &0.64 &0.82 &\textbf{51.03}\\
   &FLTrust    &94.51  &6.15   &7.32  &8.21  &9.05   &3.93   &9.93  &\textbf{51.28}\\
&FoolsGold   &93.67   &5.83   &8.62  &1.47   &6.53   &4.26   &5.73  &\textbf{50.67}\\

    \midrule
     \multirow{8}{*}{\centering \begin{tabular}{c}CIFAR10\\(Conv8)\end{tabular}}
    &FRL          & 85.30     & 12.10     & 47.25     & 12.20     & 49.62     & 34.79     & 14.42     & \textbf{71.65} \\
    &Multi-krum   & 75.02     & 18.61     & 39.47     & 0.00      & 38.74     & 32.00     & 7.14      & \textbf{68.39} \\
    &AFA          & 84.07     & 25.87     & 46.29     & 11.65     & 48.96     & 40.53     & 16.72     & \textbf{71.59} \\
    &Fang  & 84.45     & 0.39      & 35.67     & 9.96      & 37.68     & 24.91     & 30.05     & \textbf{70.41} \\
     &FABA &80.00 &19.45 &31.11 &8.74 &33.89 &31.39 &19.83 &\textbf{67.80} \\
    &DnC &81.01 &16.51 &29.63 &9.44 &35.01 &25.46 &12.16 &\textbf{64.95} \\
    &FLTrust &75.38 &8.21 &23.40 &4.84 &25.13 &10.65 &5.12 &\textbf{64.85} \\
    &FoolsGold &74.89 &27.75 &33.60 &0.00 &15.92 &10.54 &3.93 &\textbf{70.92} \\
    
    \midrule
     \multirow{8}{*}{\centering \begin{tabular}{c} FashionMNIST\\(LeNet)\end{tabular}}
    &FRL         &81.08 &14.90 &31.22 &19.45 &18.70 &18.22 &9.63 &\textbf{50.77}  \\
    &Multi-krum   &80.88 &15.06 &9.00 &19.70 &15.36 &16.72 &32.15 &\textbf{49.21} \\
    &AFA         &82.89 &17.48 &12.01 &22.09 &25.03 &21.58 &28.29 &\textbf{51.29} \\
    &Fang  &82.10 &22.23 &10.00 &20.83 &19.61 &15.96 &18.67 &\textbf{40.50} \\
     &FABA  &82.63 &27.80 &9.01 &9.94 &24.75 &18.53 &14.41 &\textbf{51.40} \\
    &DnC &80.00 &6.63 &8.40 &13.94 &6.21 &8.66 &24.03 &\textbf{56.19} \\
    &FLTrust &82.34 &10.43 &16.43 &12.91 &19.94 &17.78 &19.84 &\textbf{53.23} \\
    &FoolsGold  &76.42 &11.79 &16.62 &10.53 &14.37 &12.42 &10.12 &\textbf{51.01} \\ \midrule
    \multirow{8}{*}{\centering\begin{tabular}{c}EMNIST\\(Conv2)\end{tabular}}
    &FRL &78.08	&12.44	&3.16	&11.16	&14.24	&9.48	&14.69	&\textbf{55.90}\\
    &multi-krum &78.88	&14.88	&5.54	&19.95	&17.53	&20.20	&20.50	&\textbf{50.27}\\
    &AFA &78.29	&17.09	&6.28	&22.20	&16.71	&19.11	&12.62	&\textbf{53.58}\\
    &Fang &78.12	&8.60	&8.21	&14.06	&11.25	&21.04	&17.08	&\textbf{43.44}\\
    &FABA &78.26	&16.82	&7.56	&18.85	&17.02	&15.58	&21.78	&\textbf{59.89}\\
    &DnC &78.00	&10.26	&11.28	&16.67	&8.74	&10.81	&8.73	&\textbf{54.12}\\
    &FLTrust &78.23	&14.36	&13.08	&14.36	&10.52	&11.80	&10.52	&\textbf{53.65}\\
    &FoolsGold &70.42	&7.84	&16.22	&7.70	&2.02	&3.44	&-0.82	&\textbf{53.20}\\     
    \bottomrule
  \end{tabular}%
  }
\end{table*}
In this section, we compare our attacks with state-of-the-art poisoning attacks described in Section \ref{sec: attacks}. 
Note that for the Min-max and Min-sum attacks, we follow previous work \cite{shejwalkar2021manipulating} to assume the adversary has all benign updates.
The results are given in Table \ref{tab:attack-impact}; the ‘No attack’ column shows benign accuracy $\text{acc}_{\text{benign}}$ of the global model, while the rest of the columns show the `attack impact’ $\phi$, as defined in Section \ref{sec: set_up}.

\noindent\textbf{VEM outperforms existing attacks.}
Table \ref{tab:attack-impact} shows that our attack outperforms other attacks for all the combinations of the model, AGR, and dataset, which shows an overall 53.23\% attack impact and are $3.7 \times$ more impactful than others.
Specifically, under FRL defense, our attack achieves 61.25\% attack impact, which is $3.8 \times$ more impactful than others.
In the most robust case, Fang integrated FRL, our attack still achieves 41.28\% overall attack impact and is $4 \times$ more impactful than other attacks.
In the easiest task, the MNIST dataset, our attack achieves an overall 48\% attack impact, which is $10.5 \times$ more impactful than others.
Experiment results for CIFAR100, Location30, Purchase100, and Texas100 can be seen in Appendix \ref{app:large_dataset}.

\begin{figure}
    \centering
    \includegraphics[width=0.9\columnwidth]{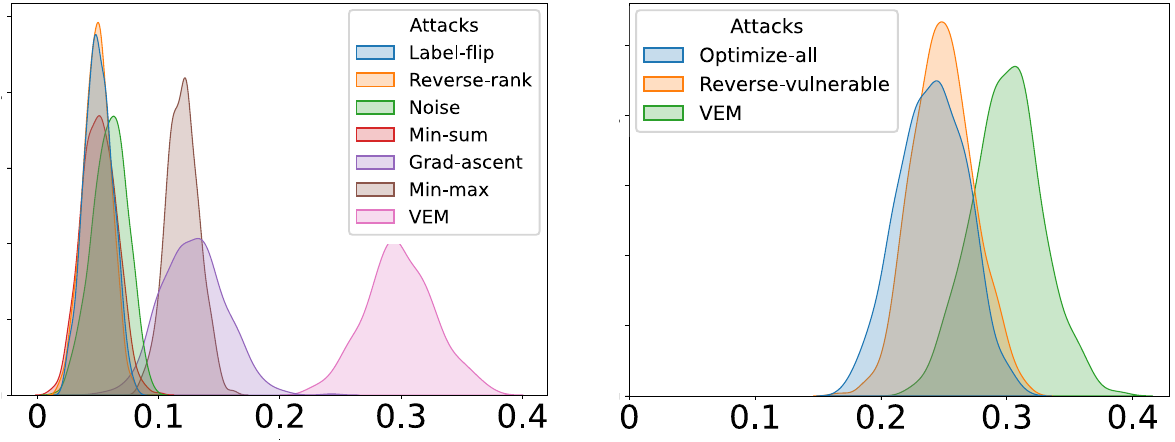}
    \caption{Edge cross rate ($\rho$) on MNIST dataset using FRL framework.}
    \label{fig:success_rate}
\end{figure}

\noindent\textbf{Why VEM outperforms existing attacks.} We summarize the reason for the success of our attack as follows. First, we identify and target the most vulnerable edges within the network. This precision ensures that our attack is more focused and effective.
Second, we employ an optimization-based approach to amplify the adversary's impact. Together, these components allow VEM to affect more edges across the selection boundary, resulting in a more successful outcome. 
As shown in  Fig. \ref{fig:success_rate} (left), our attack can achieve a 30\% edge cross rate $\rho$ while others are less than 20\%.

\subsection{Effect of Vulnerable Edge Identification and the Optimization}
\label{sec:naive}

To validate the effect of the two components, i.e., the vulnerable edge identification and the optimization mentioned in Section \ref{sec: main_result}, we design the following two attacks: 

\noindent\textbf{Design 1} (Optimize-all). We opt to maximize the \textit{aggregated reputation} difference between every edge by solving the following optimization function: 
\begin{equation}
\label{eq:optimze-all}
\begin{alignedat}{5}
& \arg\max_{P_1,P_2,\dots,P_m} \left\| {W}_m- \widetilde{{W}}_m \right\|, \\
 \text{s.t.} \
&{W}_m=A \sum_{u=1}^{m} R_u^{2d}, \\
&\widetilde{{W}}_m=A \sum_{u=1}^{m} (R_u^{2d} P_u).
\end{alignedat}
\end{equation}

\noindent\textbf{Design 2} (Reverse-vulnerable). We first identify the vulnerable edges using Theorem 1 and then simply reverse the ranking of identified vulnerable edges.

Table \ref{tab:other-two} demonstrates that our attack is $2.7 \times$ and $2.2 \times$ more impactful than the other naive designs. The reason our VEM outperforms the Optimize-all approach is due to the existence of an optimization budget. Specifically, when optimizing Eq. (\ref{eq:optimze-all}), the model tends to assign the highest rank to the lowest ranking. This behavior is similar to the `reverse-rank' attack \cite{mozaffari2023every}. As demonstrated in Theorem 1, only edges within the vulnerable range can be altered cross the selection boundary. Consequently, optimizing edges outside this range will reduce the overall damage. Additionally, in comparison with Reverse-vulnerable attack, our VEM further maximizes the \textit{reputation} difference, thus leading to a more successful attack.
As shown in Fig. \ref{fig:success_rate} (right), our attack has a higher edge cross rate.
\begin{table}[h]
  \centering
  \caption{The attack impact $\phi$ (\%) of the `Optimize-all' (without the vulnerable edge identification) design, the `Reverse-vulnerable' (without the optimization) design, and VEM under FRL and AGR integrated FRL on the MNIST dataset. We bold the strongest attack.}
  \label{tab:other-two}
  \resizebox{0.8\columnwidth}{!}{%
  \begin{tabular}{@{}c|ccc@{}}
    \toprule
    AGRs          & Optimize-all   & Reverse-vulnerable   & VEM \\ \midrule
    FRL             & 23.96           & 24.79
             & \textbf{55.80}     \\
    Multi-krum      & 22.66           & 23.80             & \textbf{48.17}     \\
    AFA             & 21.99           & 22.97             & \textbf{58.88}     \\
    Fang      & 15.01           & 18.72             & \textbf{47.70}     \\
    FABA            & 21.35           & 27.46             & \textbf{50.19}     \\
    DnC             & 11.40           &18.25            & \textbf{51.03}     \\
    FLTrust            &16.82          &  22.01           & \textbf{46.67}     \\
    FoolsGold            & 20.21           &  24.05         & \textbf{49.67}     \\
    
    \bottomrule
  \end{tabular}}
  
\end{table}
\vspace{-3mm}

\subsection{Evaluation of Estimation Methods}
\label{sec:eva_estimation}
\begin{figure}[htbp]
    \centering
    \includegraphics[width=0.95\columnwidth]{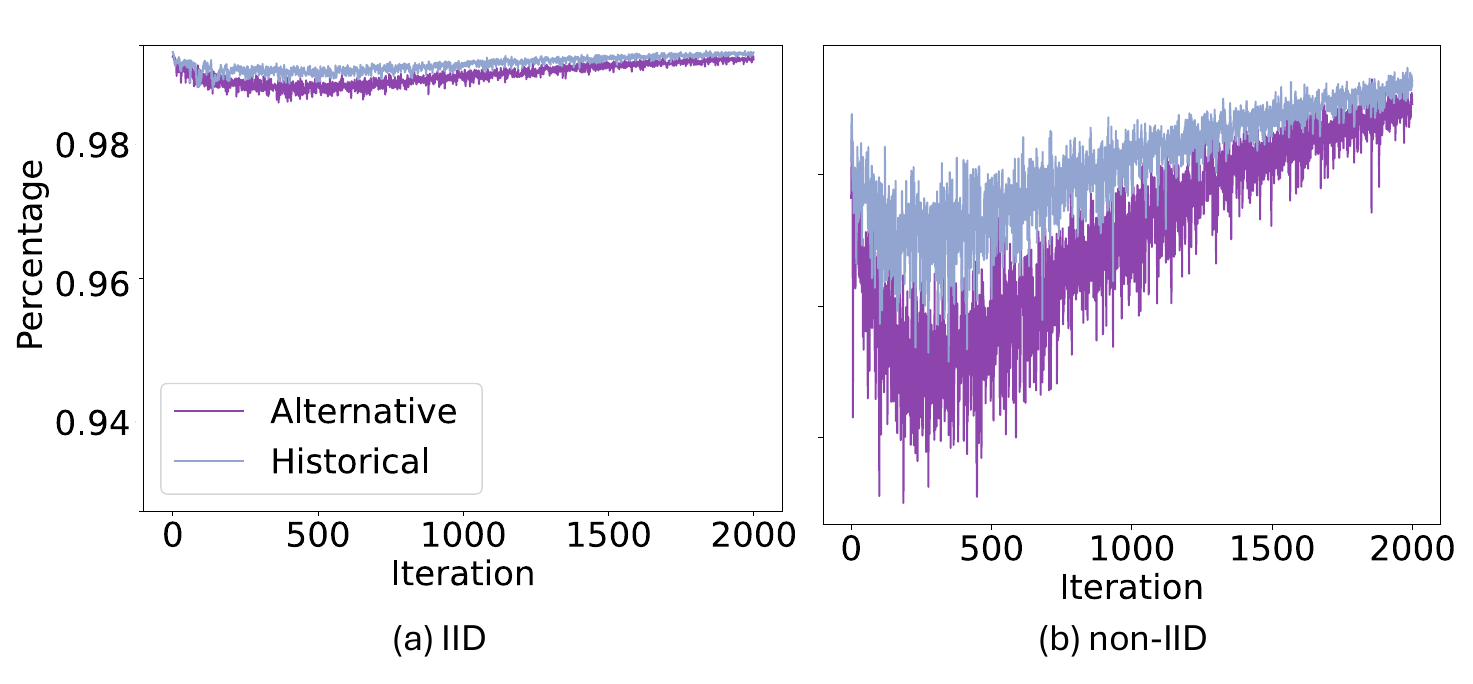}
    \caption{\textit{Estimation accuracy} during the FRL training.}
    \label{fig:estimation}
\end{figure}

We evaluate the performance of the two estimation methods based on their ability to accurately identify the true vulnerable edges.
Specifically, we first calculate the estimated $\overline{W}$  and get the vulnerable edges set using these two methods separately $\mathbb{V}_\text{alt / hist}$. We then compare these sets with the true set of vulnerable edges $\mathbb{V}_\text{true}$ which is determined using full knowledge (ideal case).
Then we calculate the \textit{estimation accuracy} (i.e., $|\mathbb{V}_\text{alt / hist} \cap \mathbb{V}_\text{true} |/|\mathbb{V}_\text{true}|$) to measure the estimation performance. 
A higher value indicates a more accurate estimation. 
As illustrated in Fig. \ref{fig:estimation}, where the y-axis represents the \textit{estimation accuracy} and the x-axis depicts training iteration.
In IID case, both methods achieve a \textit{estimation accuracy} above 98\%, and in non-IID case, the historical estimation's accuracy is above 95\%.

This high estimation accuracy for the historical method in non-IID case can be attributed primarily to two key factors:
Firstly, as shown in Theorem 1, we only need to estimate the \textbf{aggregated} reputation of benign clients to identify vulnerable edges. This aggregated result makes our method less sensitive to individual client updates. 
Secondly, although the historical estimation method relies on the assumption that the aggregated client reputation remains relatively stable across rounds. This assumption is reasonable because, in each training round, clients receive the global ranking and train for only a few epochs (e.g., 2). As a result, \( R_u(t) \) remains relatively close to \( R_g(t-1) \). 
Significant deviations would also negatively impact global model convergence. Additionally, as the model converges, this approximation becomes more accurate even in non-IID settings. 

\subsection{Impact of FRL Parameters}
\label{sec:other_exp}
In this section, we study the impact of different FRL settings on VEM, including the degree of non-IID data, the percentage of malicious clients, client participation rate, the proportion of poisoned iterations and layers, and different estimation methods. Additional experiment results, including local training epoch and hyper-parameters, are given in Appendix \ref{app: addtional_exp}.
\begin{figure*}[t]
    \centering
\includegraphics[width=\textwidth]{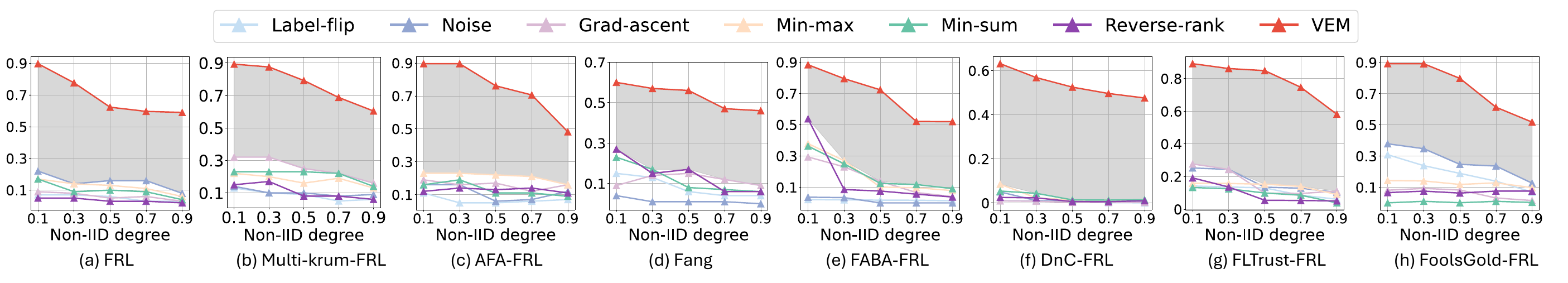}
    \caption{Attack impact (y-axis) for different attacks on different non-iid degree $\beta$ (x-axis) on MNIST dataset. A smaller $\beta$ suggests a higher non-IID degree.}
    \label{fig:non-iid}
\end{figure*}

\noindent\textbf{Impact of non-IID degree.}
We consider the attack impact of different non-IID degrees ranging from 0.1 to 1 on MNIST datasets. The hyperparameter $\beta$ dictates the extent of non-IID, where a smaller $\beta$ suggests a higher non-IID degree. We assume 20\% of the clients are malicious and analyze the effects of all model poisoning attacks.
The results depicted in Fig. \ref{fig:non-iid} demonstrate that our VEM surpasses other existing attacks in terms of effectiveness across all levels of non-IID data distribution. Notably, as the non-IID degree increases, the impact of our attack, along with the performance gap between our attack and the others, also increases. 

Here, we provide a detailed explanation of why our attack remains effective in the non-IID setting. As discussed in Section \ref{sec:eva_estimation}, our historical estimation achieves the desired performance, ensuring that the non-IID nature of the data does not significantly hinder our attack.
Moreover, our attack performs even better as the non-IID rate increases. This is because higher non-IID levels weaken the consensus among benign clients, leading to a decrease in $\overline{w}_{\text{max}}$ and an increase in $\overline{w}_{\text{min}}$ in Eq. \ref{eq:theorem}. This expansion of the vulnerable edge range reduces our attack’s dependence on specific edges, as we can effectively identify most of them. Additionally, diminished consensus among benign clients makes edges more susceptible to manipulation, further increasing the success rate of our attack.

\begin{figure*}[t]
    \centering    \includegraphics[width=\textwidth]{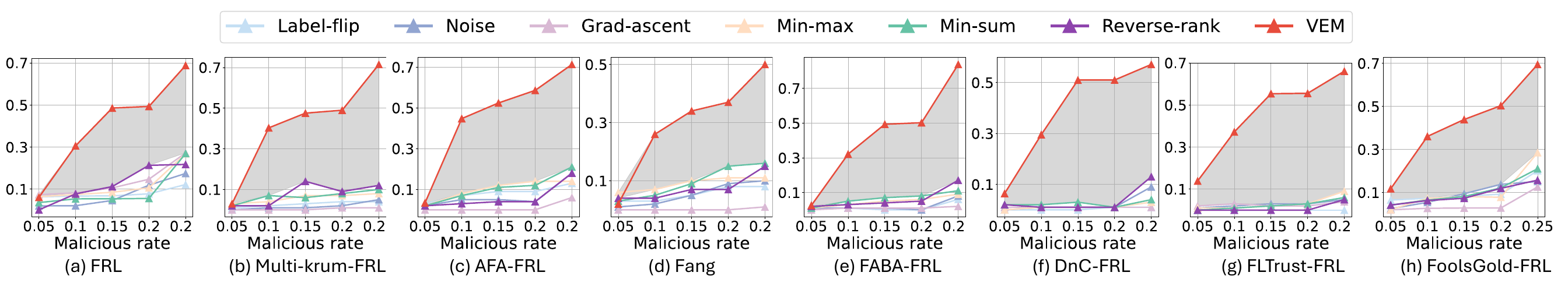}
    \caption{Attack impact (y-axis) for different attacks on different AGRs with different malicious rates (x-axis) on MNIST dataset.}
    \label{fig:malicious-rate}
\end{figure*}

\noindent\textbf{Impact of the percentage of malicious clients.}
Fig. \ref{fig:malicious-rate} shows the attack impacts of different percentages of malicious clients varied from 5\% to 25\% conducted on the MNIST dataset. We note that our attacks outperform the existing attacks by large margins in 47 out of 50 cases, with the gap ranging from 1\% to 48\%. For all of the AGRs, with an increasing percentage of malicious clients, the impacts of our attack and the gaps between our attack and existing attacks increase. 
Table. \ref{tab:mal_rate_more_dataset} shows attack impact on more datasets with malicious rate ranking from 10\% to 25\%. 

\begin{table}[h!]
\centering
\resizebox{\columnwidth}{!}{%
\begin{tabular}{c|ccccc}
\toprule
 & CIFAR10    & FashionMNIST  & Location30 & Purchase100 & Texas100 \\
  & (Conv8)  & (LeNet) &  (FC) &  (FC) &  (FC)\\ \midrule
10\%& 64.51&39.10 &88.01 & 95.65& 51.56 \\ \midrule
15\%& 66.41&42.47 &89.20 & 98.43& 52.53 \\ \midrule
20\% & 71.73 & 50.77&  90.02  &  99.80& 67.19 \\ \midrule
25\% & 81.69 & 75.84&  93.22  &  99.80& 88.21 \\
\bottomrule
\end{tabular}}
\caption{Attack impact (\%) of VEM on more datasets and model architectures with different malicious rates under FRL framework.}
\label{tab:mal_rate_more_dataset}
\end{table}
\vspace{-3mm}
\noindent\textbf{Impact of the number of clients.}
\begin{figure}[htbp]
    \centering
    \includegraphics[width=0.95\columnwidth]{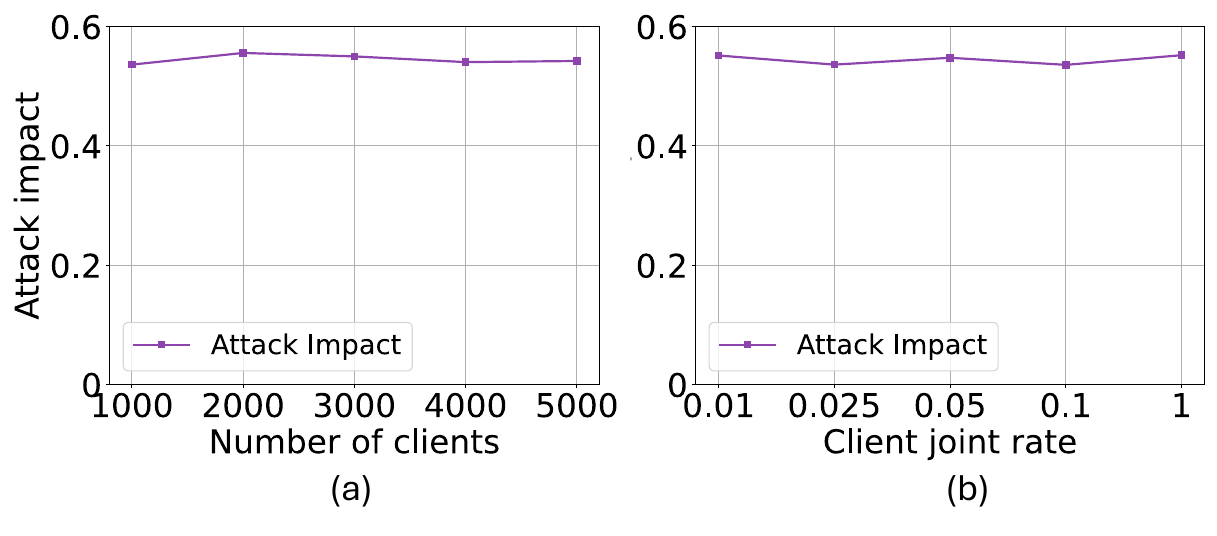}
    \captionsetup{skip=5pt}
    \caption{(a) Attack impact for the different number of clients, (b) attack impact for the different client joint rates conducted on MNIST dataset under FRL framework.}
    \label{fig:client}
\end{figure}
Fig. \ref{fig:client} (a) shows the attack impact of different numbers of total clients involved in the FL training on the MNIST dataset with FRL defense. In each round, we randomly select 2.5\% of the total clients for training.  We observe that the total number of clients has almost no impact on the effectiveness of our attack. 

\noindent\textbf{Impact of client participation rate.}
In each training round, the server selects a fraction of the clients to participate in training.
Fig. \ref{fig:client} (b) shows the attack impact of our attack when the client participation rate ranges from 0.01 to 1 on the MNIST dataset with the FRL framework. We observe that the participation rate has almost no impact on the effectiveness of our attack, which demonstrates the scalability of our attack.

\begin{figure}[htbp]
    \centering
\includegraphics[width=\columnwidth]{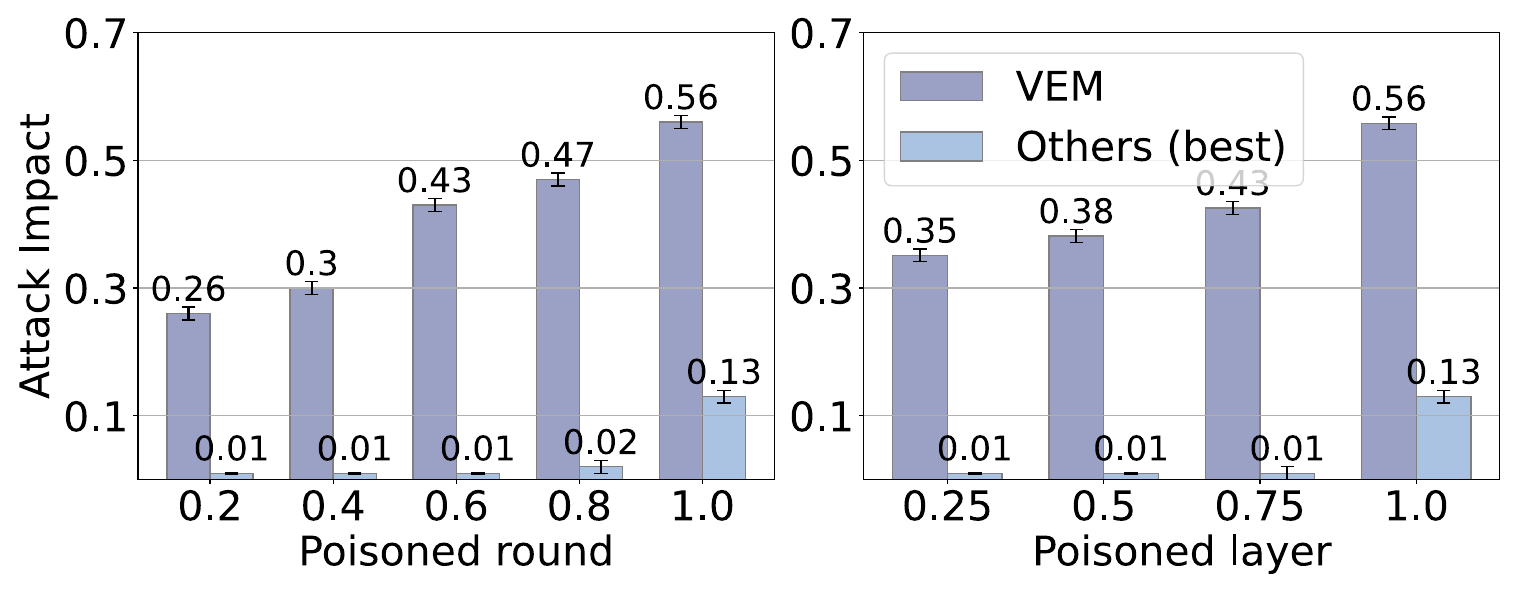}
    \caption{Attack impact for different percentages of poisoned rounds (left) on MNIST dataset under FRL framework. Attack impact for different percentages of poisoned layer (right) on MNIST dataset under FRL framework.}
    \label{fig:layer-epoch}
\end{figure}

\noindent\textbf{Impact of different percentage of poisoned iterations / layers.}
Fig. \ref{fig:layer-epoch} (left) shows the attack impact of our attacks on the MNIST dataset under FRL defense differs in the poison percentage of FL training rounds. Unsurprisingly, the attack impact increases when more rounds are poisoned. Note that our attacks achieve a 15.30\% attack impact while only using a 20\% poisoning rate, outperforming all other attacks that use a 100\% poisoning rate. Fig. \ref{fig:layer-epoch} (right) shows the attack impact of our attacks on FRL defense differs in the poison percentage of the layers. The attack impact increases with the increase in the poison layer percentage. Note that our attacks achieve a 10\% attack impact while only using a 20\% poisoning rate.

\begin{figure}[htbp]
    \centering
    \includegraphics[width=0.95\columnwidth]{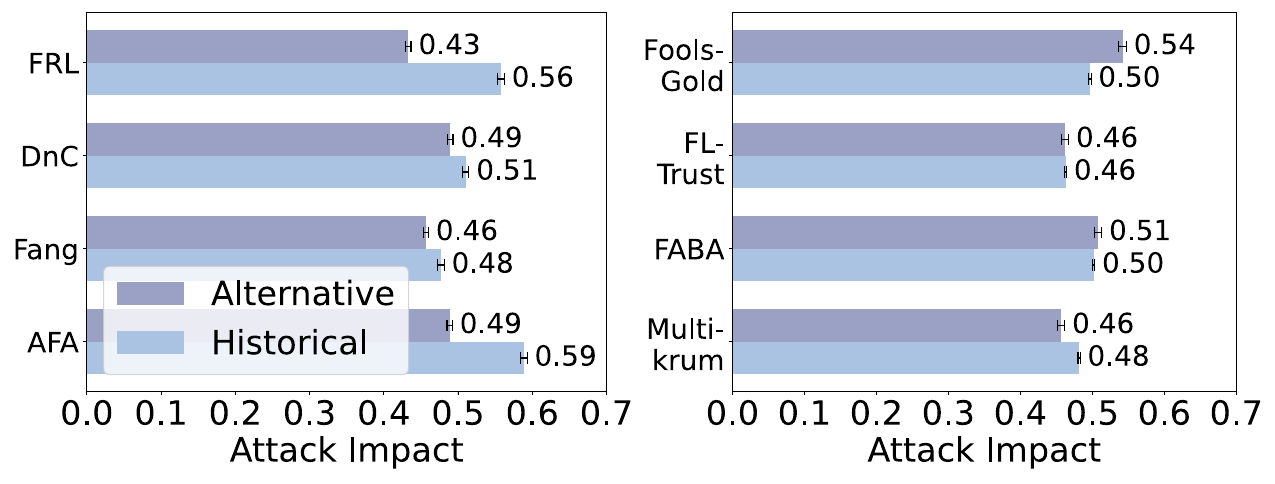}
    \caption{Attack impact for different estimation methods on MNIST dataset for FRL and AGR integrated FRL.}
    \label{fig:his-alt}
\end{figure}
\noindent\textbf{Comparison of estimation methods.}
\label{sec:estimation}
We also conduct experiments to compare the attack impact under different estimation methods discussed in Section \ref{sec: estimation}. As shown in Fig. \ref{fig:his-alt}, both of the two estimation methods achieve attack impact greater than 40\% in the MNIST dataset. Additionally, the historical estimation outperforms alternative estimation 5 out of 8 times.

\section{Discussion}
\subsection{Custom Defense}
\label{sec:defense}
As shown in Section \ref{sec:exp}, simply incorporating existing AGRs into FRL is ineffective and can sometimes have a worse effect. This is because existing AGRs primarily rely on Euclidean distance or cosine similarity to identify and exclude outliers. However, these methods do not align well with the nature of FRL, where updates do not need to be identical but should fall within a specific acceptable range. Consequently, relying solely on existing AGRs is not an optimal approach for FRL systems. With this understanding, we consider a customized defense strategy tailored to the unique requirements of FRL.


Specifically, for each user \( u \in [1, U] \), let \( \mathbb{E}_u^{\text{in}} \) denote the set of the top \( k\% \) edges selected by user \( u \).
For any two users \( u,  v  \in  [1, U] \), we define the intersection of their selected edges as
$\varphi_{u,v} = \mathbb{E}_u^{\text{in}} \cap \mathbb{E}_v^{\text{in}}$.
The size of this intersection, \( |\varphi_{u,v}| \), represents the number of edges that are common between the top \( k\% \) edges selected by users \( u \) and \( v \).
For each user \( u \), the average intersection value with all other users is computed as follows:
\begin{equation}
\label{eq:average_inter}
\mu_u = \frac{1}{U-1} \sum_{v \in [1, U], v \neq u} |\varphi_{u,v}|.
\end{equation}
Users are then clustered into two groups based on their mean intersection values \( \mu_u \) and the cluster with smaller mean values is considered potentially malicious and filtered out before majority voting, we call this defense Intersection-based defense (IBD).


\begin{table}[h!]
\centering
\resizebox{\columnwidth}{!}{%
\begin{tabular}{c|cc|cc}
\toprule
 & \multicolumn{2}{c|}{FRL} & \multicolumn{2}{c}{IBD-FRL} \\
\midrule
 & No attack & Attack impact & No attack & Attack impact \\
\midrule
FashionMNIST & 81.08 & 50.77 & 79.01 & 50.01 \\
CIFAR10 & 85.30 & 71.65 & 75.59 & 65.08 \\
\bottomrule
\end{tabular}
}
\caption{Comparison of FRL and IBD integrated FRL under VEM attack.}
\label{tab:defense}
\end{table}

\begin{figure}[!h]
    \centering
\includegraphics[width=\columnwidth]{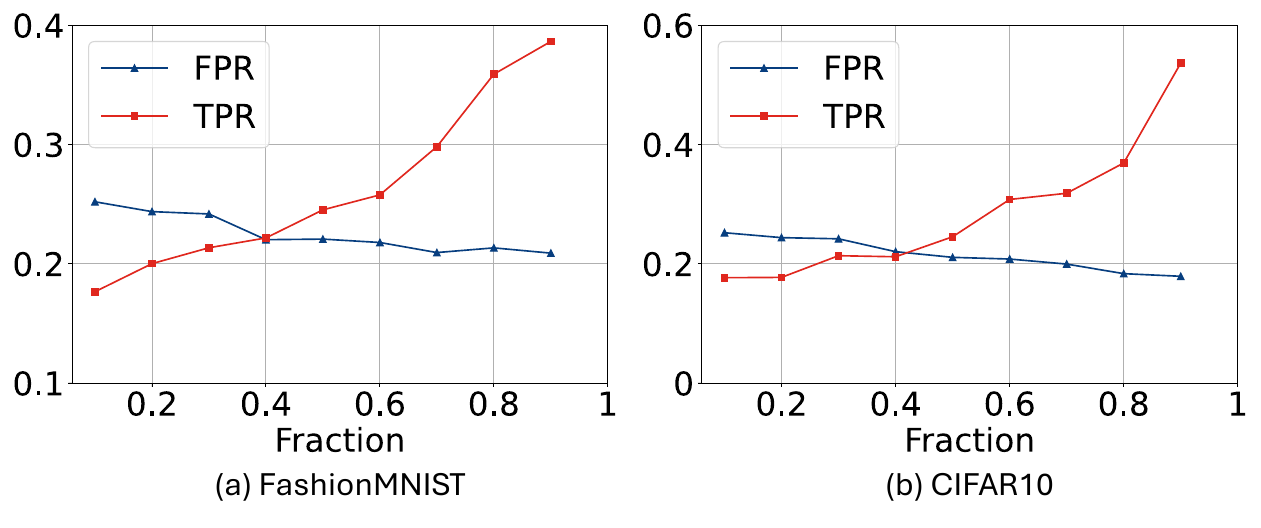}
    \caption{False positive rate (FPR) and true positive rate (TPR) of IBD defense.}
    \label{fig:fpr-tpr}
\end{figure}

\noindent\textbf{VEM remains effective against IBD.} As demonstrated in Table \ref{tab:defense}, our VEM attack continues to achieve strong attack impact despite the implementation of IBD defenses. 
This is because the identified vulnerable edges are inherently inconsistent among benign clients, resulting in minimal variation in the average intersection value $\mu_u$ for $u \in U$.
Additionally, by simply adjusting $\zeta$ in Eq.~\ref{eq:with-zeta}, we can render the malicious updates indistinguishable from benign updates.
Consequently, when the IBD defense is applied, not only does it fail to filter out the malicious updates effectively, but it also inadvertently excludes a significant number of benign clients. This will amplify the overall impact of the VEM attack. 
We also evaluate the False Positive Rate (FPR) and True Positive Rate (TPR) of the IBD defense under different values of hyper-parameter $\zeta$ in Eq. (\ref{eq:with-zeta}). As shown in Fig. \ref{fig:fpr-tpr}, as the value of $\zeta$ increases, the FPR decreases while the TPR increases. However, it still has a relatively high FPR, e.g., more than 20\% for the Fashion-MNIST dataset and around 20\% for the CIFAR10 dataset,
indicating that some benign updates are filtered out by the defense. Therefore, the IBD-integrated FRL is not yet sufficiently effective against our VEM attack.
\label{app:more_defense_exp}

Furthermore, we argue that the IBD is not an applaudable defense, as it will compromise the model performance under the no-attack scenario, leading to a 3\% degradation for the FashionMNIST dataset and 11\% degradation for the CIFAR10 dataset.
Furthermore, it also incurs significant computational overhead. Specifically, IBD adds an additional time complexity of $\mathcal{O}\left(U^2 \cdot n \cdot L\right)$ to FRL, where $U$ is the number of clients selected in the current round, $n$ is the number of parameters in each layer and $L$ is the number of layers in the model. 
We record the computational time, which reveals that IBD-FRL takes 3 $\times$ longer for the Fashion-MNIST dataset and 10 $\times$ longer for the CIFAR10 dataset.

\subsection{Limitations and Future Work}
The scope of the current work is confined to untargeted poisoning attacks, which aim to degrade the overall model performance. Future research could delve into targeted poisoning attacks, where the adversarial manipulation is aimed specifically at certain outcomes or models, providing a more nuanced understanding of how FRL systems can be compromised.  Additionally, while the existing estimation method for identifying benign updates within the FRL framework is effective in the IID setting, it remains constrained in non-IID settings. Future research could focus on augmenting these methods with advanced prediction-based techniques, such as those leveraging machine learning or statistical forecasting. 

Moreover, the development of communication-efficient and robust FL systems is essential. Previously, the non-optimizable nature of discrete spaces has posed a significant challenge to achieving such efficiency. Our work makes a crucial contribution by enabling optimization within these discrete spaces. This advancement not only provides practical strategies for overcoming a major barrier in FL system design but also lays the groundwork for deeper exploration into creating more resilient and efficient frameworks. By offering a method to navigate and optimize within discrete spaces, our research opens new pathways for enhancing both the robustness and communication effectiveness of FL systems. These insights can guide future studies and innovations aimed at building FL architectures that can maintain strong performance even under constrained communication conditions.

Moreover, future research could explore certified robustness metrics.
One possible evaluation approach is to determine the maximum number of malicious clients that the FRL system can tolerate under our attack. 
This requires further investigation into how changes in selected edges impact
the model prediction results.
Exploring this area could provide valuable insights for developing more robust FL systems.




\section{Conclusion}
In this work, we present the first comprehensive evaluation of the robustness of the promising ranking-based FL system.
Our evaluation uncovers a significant vulnerability within this system, that certain edges are particularly susceptible to poisoning attacks. Through a theoretical investigation, we demonstrate the existence of these vulnerable edges and establish both lower and upper bounds to identify them within each layer. By exploiting these vulnerabilities, our VEM can strategically target and manipulate these weak points. Extensive experimental results show that the VEM attack significantly outperforms existing state-of-the-art attacks across various network architectures and datasets. Moreover, our VEM attack remains effective even when faced with customized defense mechanisms, underscoring its resilience and the importance of addressing these vulnerabilities in future FL system designs. 
Furthermore, our work offers insights into how optimization in a discrete space can be effectively approached, providing guidance for developing more robust and communication-efficient FL systems. 




%
\bibliographystyle{IEEEtran}
\bibliography{main}

\begin{appendices}
\section{Algorithms and Proof}
\subsection{Algorithms}
\label{app:alg}
\vspace{-10pt}

\begin{algorithm}[H]
\caption{Federated Ranking Learning (FRL)}
\label{alg:frl}
\begin{algorithmic}[1]
\State \textbf{Input:} Number of rounds $T$, number of local epochs $E$, number of users per round $U$, seed SEED, learning rate $\eta_\text{ep}$, subnetwork size $k\%$
\State \textbf{Server: Initialization}
\State $\theta^s, \theta^w \gets \text{Initialize scores and weights using } \text{SEED}$
\State $R_g \gets \text{ARGSORT}(\theta^s)$ \Comment{Sort the initial scores and obtain initial rankings} 
\For{$t \in [1, T]$}
    \State $U \gets \text{Select $U$  clients from $N$ total clients}$ 
    \For{$u \in [1, U]$}
        \State {// Clients: calculate the rankings}
        \State $\theta^s, \theta^w \gets \text{Initialize scores and weights using } \text{SEED}$
        \State $\theta^s[R^t_g] \gets \text{SORT}(\theta^s)$ \Comment{Sort scores based on the global ranking}
        \State $\theta^s \gets \text{Edge-PopUp}(E, D^{tr}_u, \theta^w, \theta^s, k, \eta_\text{ep})$ \Comment{Client $u$ uses Edge-PopUp to train a supermask on its local training dataset}
        \State $R^t_u \gets \text{ARGSORT}(\theta^s)$ \Comment{Ranking of the client}
    \EndFor
    \State {// Server: majority voting}
    \State $R^{t+1}_g \gets \text{VOTE}(R^t_{\{u \in [1,U]\}})$ 
\EndFor
\Function{VOTE}{$R_{\{u \in U\}}$}
    \State $V \gets \text{SUM}(\text{ARGSORT}(R_{\{u \in [1, U]\}}))$, $A \gets \text{SUM}(V)$
    \State \textbf{return} ARGSORT$(A)$
\EndFunction
\end{algorithmic}
\end{algorithm}

\subsection{Proof}
\label{proof:theorm1}

In this section, we detail the proof of Theorem 1. 
We will prove the upper bound of $\overline{w}_v$, and the method of proving the lower bound will be similar. 



Let $e_v$ denote an edges that is selected by $U-m$ clients before the attack, such that $e_v \in \overline{\mathbb{E}}^{\text{in}}$, then we have $\overline{w}_{\text{min}} \leq \overline{w}_v$. Since
\begin{equation}
\label{eq:calculate_w}
    \overline{w}_j=\sum_{i=1}^{n} a_i \sum_{u=m+1}^{U} R_u ^{2d} [i,j],
\end{equation}
then we have
\begin{equation}
\label{eq: before_attack}
\begin{alignedat}{2}
   &\sum_{i=1}^{n} a_i \sum_{u=m+1}^{U} R_u ^{2d} [i,\text{min}]  \leq & \sum_{i=1}^{n} a_i \sum_{u=m+1}^{U} R_u ^{2d} [i,v].
\end{alignedat}
\end{equation}
Assume that $e_v$ is a vulnerable edge, such that $e_v \in \widetilde{\mathbb{E}}^{\text{out}}$ after the attack, therefore we have $\widetilde{w}_v < \widetilde{w}_{\text{min}}$. Since
\begin{equation}
    \widetilde{w}_j=\sum_{i=1}^{n} a_i \sum_{u=1}^{m} \widetilde{R}_u ^{2d} [i,j]+\sum_{i=1}^{n} a_i \sum_{u=m+1}^{U} R_u ^{2d} [i,j],
\end{equation}
then we have
\begin{equation}
\label{eq: after_attack}
\begin{alignedat}{2}
   &\sum_{i=1}^{n} a_i \sum_{u=1}^{m} \widetilde{R}_u ^{2d} [i,v]+\sum_{i=1}^{n} a_i \sum_{u=m+1}^{U} R_u ^{2d} [i,v] \\ < & \sum_{i=1}^{n} a_i \sum_{u=1}^{m} \widetilde{R}_u ^{2d} [i,\text{min}]+\sum_{i=1}^{n} a_i \sum_{u=m+1}^{U} R_u ^{2d} [i,\text{min}].
\end{alignedat}
\end{equation}
To achieve Eq. (\ref{eq: after_attack}) given the condition of Eq. (\ref{eq: before_attack}), the extreme case is to find $\widetilde{R}_{u} ^{2d} [i,v]$ and $ \widetilde{R}_{u} ^{2d} [i,\text{min}] $ for $u \in m$ that can make  
\begin{equation}
\label{eq:min}
      \sum_{i=1}^{n} a_i \sum_{u=1}^{m} \widetilde{R}_u ^{2d} [i,v]-\sum_{i=1}^{n} a_i \sum_{u=1}^{m} \widetilde{R}_u ^{2d} [i,\text{min}]
\end{equation}
as small as possible. 
Since $A$ is an increasing vector and $R_{u}^{2d}$ (for $u \in m$) are permutation matrics, the extreme case in Eq. (\ref{eq:min}) is 

\begin{equation}
\label{eq:solution}
\begin{alignedat}{2}
\widetilde{R}_u^{2d}[i,v] &= \begin{cases}
1 & \text{if } i = 1, \\
0 & \text{otherwise},
\end{cases} \\
\text{and} \
\widetilde{R}_u^{2d}[i,\text{min}] &= \begin{cases}
1 & \text{if } i = n, \\
0 & \text{otherwise}. \\
\end{cases}
\end{alignedat}
\end{equation}
Expanding Eq. (\ref{eq: after_attack}), we have
\begin{equation}
\label{eq:long}
\begin{alignedat}{3}
 &\sum_{i=2}^{n} a_i \sum_{u=1}^{m} \widetilde{R}_u ^{2d} [i,v] + a_1 \sum_{u=1}^{m} \widetilde{R}_u ^{2d} [1,v] \\
 &+\sum_{i=1}^{n} a_i \sum_{u=m+1}^{U} R_u ^{2d} [i,v] \\
<& \sum_{i=1}^{n-1} a_i \sum_{u=1}^{m} \widetilde{R}_u ^{2d} [i,\text{min}]+a_n \sum_{u=1}^{m} \widetilde{R}_u ^{2d} [n,\text{min}] \\
&+\sum_{i=1}^{n} a_i \sum_{u=m+1}^{U} R_u ^{2d} [i,\text{min}].
\end{alignedat}
\end{equation}
Since $\widetilde{R}_u^{2d}[i,v]=0,\ \text{where} \ i \in [2,n]$ and $\widetilde{R}_u^{2d}[i,\text{min}]=0,\ \text{where} \ i \in [1,n-1]$, Eq. (\ref{eq:long}) can be further simplify as follows:
\begin{equation}
\label{eq:simplify}
\begin{alignedat}{3}
 & a_1 \sum_{u=1}^{m} \widetilde{R}_u ^{2d} [1,v]
 +\sum_{i=1}^{n} a_i \sum_{u=m+1}^{U} R_u ^{2d} [i,v] \\
 <& \ a_n \sum_{u=1}^{m} \widetilde{R}_u ^{2d} [n,\text{min}] +\sum_{i=1}^{n} a_i \sum_{u=m+1}^{U} R_u ^{2d} [i,\text{min}].\\
\end{alignedat}
\end{equation}
Combining Eq. (\ref{eq:calculate_w}), Eq. (\ref{eq:solution}) and Eq. (\ref{eq:simplify}), we have
\begin{equation}
\label{eq:final}
\begin{alignedat}{3}
 \overline{w}_v < \overline{w}_{\text{min}}+ m(a_n-a_1).
\end{alignedat}
\end{equation}

\section{Additional Details on Experimental Setup}
\label{sec:appdex_exp}
\subsection{Detailed Settings}
\label{app:fig3}
For Fig. 3, we conduct the experiment on MNIST (Conv2) and CIFAR-10 (Conv8) using the FRL framework with $k=50\%$ in a benign setting. In each round, we randomly select 25 clients for training and run the experiment for 1000 rounds. During each training round, we record the global ranking $R_g$ to calculate the 	\textit{reputation} difference between each edge and the selection boundary. We then calculate the mean value of the reputation difference for each edge across 1000 rounds and use a bar graph (Fig. 3) to show its distribution.
Additionally, we add a red dashed vertical line in Fig. 3 to indicate the `max adversary damage'. As the experiment is conducted in a benign setting, the `max adversary damage' is calculated manually using $m(a_n - a_1)$, where we assume $m = 20\%$.
\subsection{Datasets and Model Architectures}
\label{sec:detailed_dataset}

\noindent\textbf{MNIST} \cite{lecun1998mnist}: We Uses Conv2 architecture with 10\% of parameters for sparsity. Each client trains for 2 epochs with SGD (LR=0.4, momentum=0.9, weight decay=1e-4, batch size=32).
\\
\textbf{CIFAR-10} \cite{krizhevsky2009learning}: We ues Conv8 architecture with 50\% of parameters for sparsity. Each client trains for 5 epochs with SGD (same settings as MNIST).
\\
\textbf{Fashion-MNIST} \cite{xiao2017fashion}
We uses LeNet architecture with 50\% of parameters for sparsity. Each client trains for 2 epochs with SGD (same settings as MNIST).
\\
\textbf{EMNIST} \cite{cohen2017emnist}: We uses LeNet architecture with 50\% of parameters for sparsity. Each client trains for 2 epochs with SGD (same settings as MNIST).
\\

\subsection{Evaluated Attacks and Settings}
\label{sec:detailed_attack}
\noindent\textbf{Label-flip attack} is a widely referenced data poisoning attack where the adversary intentionally switches the label for training data. Following the setting as \cite{biggio2012poisoning}, we flip the label $y$ to $C - y - 1$, where $C$ denotes the number of classes in the dataset.

\noindent\textbf{Noise attack} randomly adds noise on the gradient of the compromised worker devices. Here, we add random noise to the score of each compromised worker to make it update malicious rankings.

\noindent\textbf{Grad-ascent attack} \cite{kurakin2016adversarial} is to revise the gradient towards the adversarial direction. Here, we flip the sign of each dimension of the score to make it generate malicious rankings. 

\noindent\textbf{Min-max attack} \cite{shejwalkar2021manipulating} is the optimization-based work of model poisoning attacks, which computes the malicious gradient such that its maximum distance from any other gradient is upper bounded by the maximum distance between any two benign gradients. Hence, they try to solve $\arg\max_\gamma \max_{i \in [n]} \|\nabla_m - \nabla_i\|_2 \leq \max_{i,j \in [n]} \|\nabla_i - \nabla_j\|_2$, s.t. $\nabla_m = \text{f}_{\text{avg}}(\nabla_{\{i \in [n]\}}) + \gamma\nabla_p$. 

\noindent\textbf{Min-sum attack} \cite{shejwalkar2021manipulating} ensures that the sum of squared distances of the malicious gradient from all the benign gradients is upper bounded by the sum of squared distances of any benign gradient from the other benign gradients. Hence, they try to solve $\arg\max_\gamma \sum_{i \in [n]} \|\nabla_m - \nabla_i\|_2^2 \leq \max_{i \in [n]} \sum_{j \in [n]} \|\nabla_i - \nabla_j\|_2^2 $ s.t. $\nabla_m = \text{f}_{\text{avg}}(\nabla_{\{i \in [n]\}}) + \gamma\nabla_p$.

\noindent\textbf{Reverse-rank} \cite{mozaffari2023every} is the SOTA untargeted poisoning attack on FRL. In this attack, malicious clients initially compute rankings on their benign data and aggregate these rankings. Subsequently, each malicious client reverses the aggregate rankings and shares them with the FRL server in a given round.
\subsection{Evaluated Defenses and Settings}
\label{app:detailed_agr}
    \noindent\textbf {Krum / Multi-Krum} \cite{blanchard2017machine} selects one or more local updates that are close to its $U - m - 2$ neighboring update in Euclidean distance as the global model; here, $m$ is an upper bound on the number of malicious clients and U is the number of clients selected for the current training. 

    
    \noindent\textbf {Adaptive federated average (AFA)} \cite{munoz2019byzantine} compares the cosine similarity between each user update and a benign gradient and discards the gradients whose similarities are out of a range; this range is a simple function of the mean, median, and standard deviation of the similarities.
    
    \noindent\textbf {FABA} \cite{xia2019faba} iteratively excludes the local updates that are furthest from the average update until the number of excluded updates matches the number of attackers. 
    
    \noindent\textbf {Divide-and-conquer} (DnC) \cite{shejwalkar2021manipulating} employs singular value decomposition (SVD) to detect and remove outliers.  Specifically, it first randomly selects a set of indices from the input gradient's dimensions to create a subsample of gradients. It then centers these subsampled gradients by subtracting their mean. Then, it projects the centered gradients onto the top singular eigenvector, which captures the most significant direction of variation in the centered gradients. Then, gradients with the highest scores are removed, and the remaining ones are considered benign. 
    
    \noindent\textbf {Trimmed Mean / Median} \cite{yin2018byzantine} employ approach involves coordinate-wise aggregation, where the AGRs separately identify and remove $m$ largest and smallest values in each dimension and aggregate the remaining values for each dimension. 
    
    \noindent\textbf {Bulyan} \cite{guerraoui2018hidden} first selects $\theta \ (\theta \leq n - 2m)$ gradients in the same fashion as Multi-krum and then computes the Trimmed-mean of the selected gradients for dimensional-wise aggregation.
    
    \noindent\textbf  {Norm-bounding} \cite{sun2019can} constrains the $L_2$ norm of all client updates to a fixed threshold before the average aggregation. For a local update $\nabla$ and threshold $T$, if $\|\nabla \|_2 > T$, $\nabla $ will be scaled by $\frac{T}{\|\nabla \|_2}$. This scaling operation effectively reduces the severity of the poisoning level of the malicious model updates. 
    
    \noindent\textbf {Centered Clip (CC)} \cite{karimireddy2021learning} prunes local updates that are too large because an attacker can upload such updates to control the global model.
    
    \noindent\textbf  {Fang} \cite{fang2020local} assumes the server holds a clean validation dataset, and the server can calculate losses and errors on the validation set of the updated model to determine whether the updates originate from malicious or benign clients.
    
    \noindent\textbf {FLTrust} \cite{caofltrust} leverages an additional validation dataset on the server. In particular, each client is given a trust score based on the distance between the local updates and the reference gradients produced from a clean, trustworthy dataset. The trust score then serves as the weight when averaging updates. 
    
    \noindent\textbf {FoolsGold} \cite{fung2020limitations} finds clients with similar cosine similarity to be malicious, penalize their reputation, and return a weighted mean of the update weighed by their reputation.

\section{Additional Experiment Results}
\label{app: addtional_exp}


\noindent\textbf{{Attack Impact on More Datasets.}}
\label{app:large_dataset}
Table \ref{tab:large-dataset} presents the impact of our attack on larger datasets, as well as three tabular datasets with 10\% and 20\% malicious rates under the FRL framework. From the table, we can see that our VEM achieves desirable attack impact in all cases. 
\begin{table}[h!]
\centering
\resizebox{\columnwidth}{!}{%
\begin{tabular}{c|ccccc}
\toprule
Malicious & CIFAR10    & CIFAR100  & Location30 & Purchase100 & Texas100 \\
 rate  & (VGG)  & (Conv8) &  (FC) &  (FC) &  (FC)\\ \midrule
10\%& 44.21&40.54 &88.01 & 95.65& 51.56  \\ \midrule
20\% & 54.23 & 47.30&  90.02  &  99.80&  67.19\\ 
\bottomrule
\end{tabular}}
\caption{Attack impact (\%) of VEM on other datasets and model architectures with different malicious rates under FRL framework.}
\label{tab:large-dataset}
\end{table}

\begin{figure}
    \centering
    \includegraphics[width=\columnwidth]{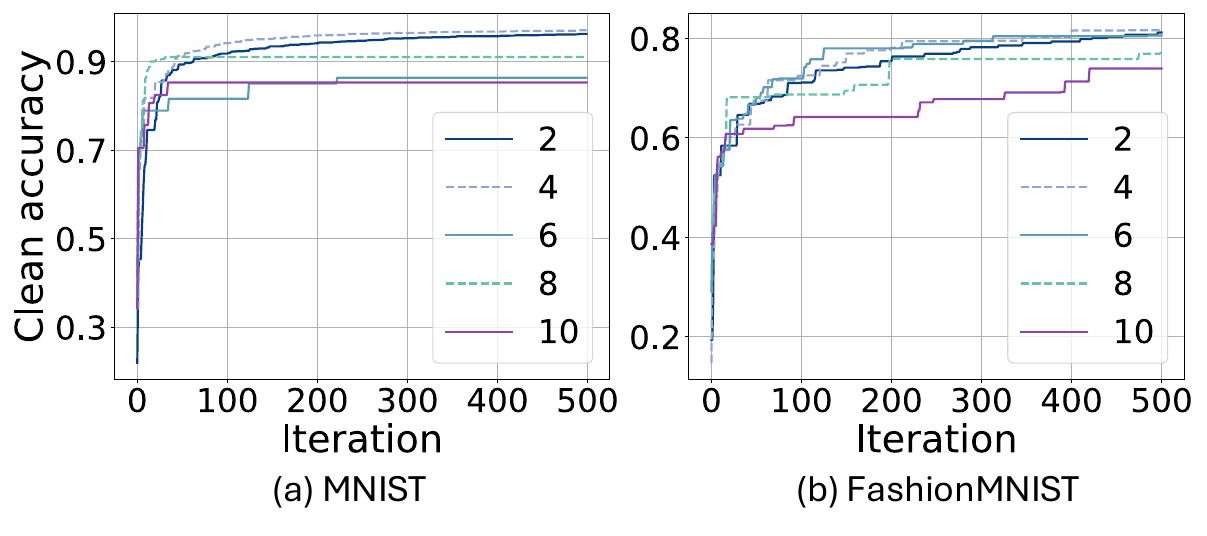}
    \caption{Clean accuracy for different local training epochs under FRL defense.}
    \label{fig:clean_local}
\end{figure}

\noindent{\textbf{Impact of Local Training Epoch.}}
Table \ref{tab:local_epoch} illustrates the attack impact of varying numbers of local training epochs, ranging from 2 to 10, on the MNIST dataset with FRL defense. We observe that as the number of local epochs increases, the attack impact slightly decreases. This is due to the reduction in prediction accuracy with more local epochs. However, the benign accuracy also declines, indicating that an increase in local training epochs negatively affects convergence ability. As shown in Fig \ref{fig:clean_local}, this phenomenon also appears in other datasets.
\begin{table}[h!]
\centering
  \resizebox{\columnwidth}{!}{%
  \begin{tabular}{c|ccccc}
  \toprule
    {Local epoch} & {2} & {4} & {6} & {8} & {10} \\ \midrule
    {Benign accuracy} & {98.63} & {97.67} & {89.38} & {86.12} & {85.32} \\   
    {Attack impact} & 56.53 & 54.55  &50.56  & 57.78 & 57.90   \\ \bottomrule
  \end{tabular}}
\caption{Attack impact for the local training epoch conducted on MNIST dataset under FRL defense.}
\label{tab:local_epoch}
\end{table}

\noindent{\textbf{Impact of Hyper-parameters.}}
\begin{figure}[!h]
    \centering
\includegraphics[width=\columnwidth]{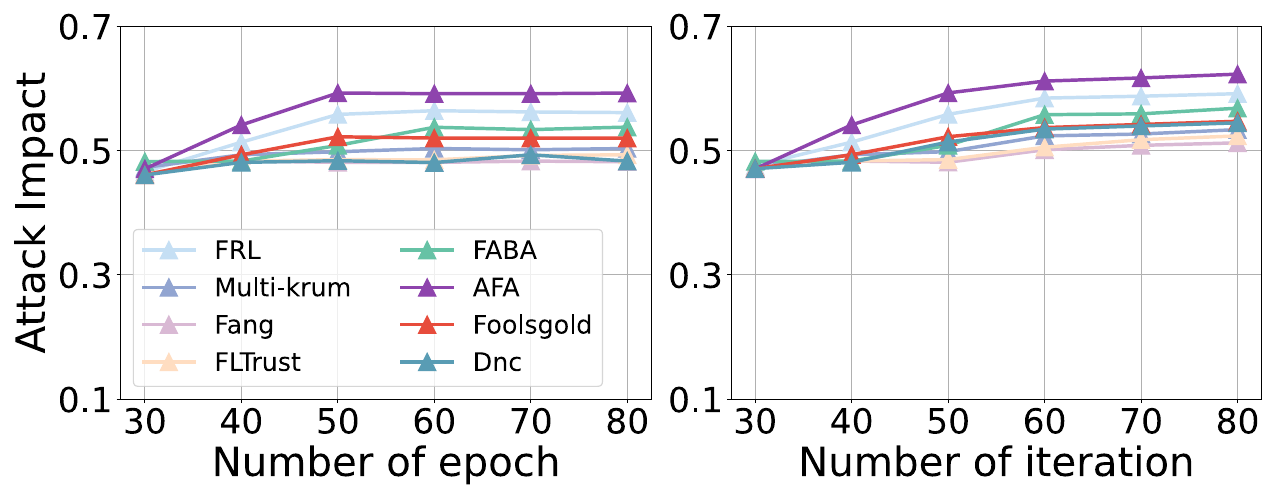}
    \caption{Effect of hyper-parameters conducted on MNIST dataset.}
    \label{fig:hyper}
\end{figure}
We also examine the effects of two hyperparameters: the \textit{training epoch} of optimization function in Eq. (\ref{eq:object-3}) and the \textit{number of iterations} in Eq. (\ref{eq:Sinkhorn}).
As illustrated in Fig. \ref{fig:hyper} (left), it depicts the attack impact across different \textit{training epochs}. The impact rises before 50 epochs and stabilizes afterward, suggesting near convergence. Consequently, we use 50 epochs for attack training in our main experiment. 
Fig. \ref{fig:hyper} (right), increasing the \textit{number of iterations}  enhances the attack impact but also raises computational costs. Therefore, we determined that 50 iterations provide the best balance for our main experiment.


\noindent{\textbf{More Defense Discussion.}}
\label{app:other_defense}
Augmented defenses perform pre-processing steps before applying existing AGRs, enhancing the robustness of the FL system. We test our attack against the SOTA augmented defenses method: FoundationFL \cite{fang2025we}.
The experiment are conducted with 10\% and 20\% malicious rates, resulting in attack impacts of 66.25\% and 72.50\%, respectively.
The reason FoundationFL fails to defend against our attack lies in its `synthetic updates generation' process. This approach first selects updates that deviate significantly from extreme updates (i.e., those with large magnitudes) and uses them as synthetic updates. However, due to the nature of FRL, the server cannot accurately identify extreme updates. Consequently, the selection process becomes essentially random, resulting in the use of arbitrary updates as synthetic ones. This effect is functionally similar to increasing the number of selected clients per round. As shown in Fig. 11 in our manuscript, our attack remains unaffected by the client participation rate, enabling it to successfully bypass this defense.

Certified defenses provide provable security guarantees against poisoning attacks, making them more secure against advanced adversarial strategies. To evaluate the impact of our attack, we test it under FLCert \cite{cao2022flcert}. 
We evaluate our attack against the FLCert defense on the CIFAR-10 dataset with 10\% and 20\% malicious rates, resulting in attack impacts of 83.33\% and 86.53\%, respectively. This outcome can be attributed to two key factors. First, in the early stages of model training, the model is not yet fully trained, leading to poor performance across all groups on the test samples. As training progresses, our attack gradually manipulates the global model, which subsequently influences each local model. As a result, the local models deviate from their optimal training trajectories, ultimately struggling on the test dataset.

\end{appendices}

\end{document}